\documentclass{article}

\usepackage[preprint]{neurips_2025}

\usepackage[utf8]{inputenc}
\usepackage[T1]{fontenc}
\usepackage{booktabs}
\usepackage{amsmath}
\usepackage{amssymb}
\usepackage{graphicx}
\usepackage{subcaption}
\usepackage{algorithm2e}
\usepackage[colorlinks=true,linkcolor=blue,citecolor=blue,urlcolor=blue]{hyperref}

\RestyleAlgo{ruled}
\SetKwComment{Comment}{/* }{ */}

\newcommand{\sptgnn}{SpTGNN}
\newcommand{\soc}{\text{SOC}}

\newcommand{\moe}{MoE}
\newcommand{\rgat}{RGAT}

\title{Multi-Modal Spatio-Temporal Graph Neural Network with Mixture of Experts
for Soil Organic Carbon Prediction}

\author{%
  Daniele Mos \quad Felipe Drummond \quad Anton Bossenbroek \quad
  Soufiane el Khinifri \\[2pt]
  Spatialise B.V. \\
  \texttt{\{daniele,\,felipe,\,anton,\,souf\}@spatiali.se}
}

\hypersetup{
  pdftitle={Multi-Modal Spatio-Temporal Graph Neural Network with Mixture of
    Experts for Soil Organic Carbon Prediction},
  pdfauthor={Daniele Mos, Felipe Drummond, Anton Bossenbroek, Soufiane el Khinifri},
}

\begin{document}

\maketitle

\begin{abstract}
\textbf{Top-soil organic carbon (SOC)} prediction for agricultural land is
fundamental to advancing agricultural sustainability, informing land use policy
and improving fertilization plans. Existing approaches face two main
limitations: \textit{first}, they employ hand-crafted environmental covariates
paired with classical machine learning or single-modal deep learning models that
do not capture rich spectral and temporal information, and \textit{second}
grid-based architectures ignore the irregular spatial structure of field
measurements.
 We introduce \sptgnn{}, a \textbf{multi-modal spatio-temporal graph neural
 network} that addresses both limitations. \textbf{\sptgnn{}} represents soil
 measurements as nodes in a heterogeneous graph with three edge types, encoding
 spatial proximity, spectral similarity and elevation similarity. The model
 applies relational graph attention to learn separate patterns per relation. A
 fine-tuned \textbf{TerraMind} \citep{jakubik2025terramind} encoder extracts
 node features from remote sensing signals collected by the Sentinel-2 and
 Sentinel-1 ESA missions and Digital Elevation Models, which \sptgnn{} combines
 with per-sample environmental covariates (climate, terrain, soil texture,
 land-use variables) and learned positional and temporal embeddings. A sparse
 \textbf{Mixture-of-Experts} module fuses the four feature streams (imagery,
 environmental, positional and temporal) via top-$k$ routing. The framework
 captures predictive uncertainty by pairing heteroscedastic regression for
 aleatoric noise with deep ensembles for epistemic variance. In addition,
 Moran's~$I$ penalty further acts as a spatial regularizer, ensuring spatial
 autocorrelation consistency. We evaluate \sptgnn{} on a global SOC corpus split
 into three regional instances ($\sim$49\,k samples globally, Africa
 $\sim$26\,k and Europe $\sim$14\,k). Our headline $5$-member deep ensemble
 \sptgnn{} reports $R^{2}=0.762$, RMSE\,$=3.51\pm0.48\,\mathrm{g\,kg^{-1}}$ and
 MAPE\,$=22.9\,\%$ on the Africa test split, a substantial improvement over a
 tabular \textit{XGBoost} baseline. The best single-model checkpoint reaches a
 validation $R^{2}=0.864$ and MAPE\,$=13.7\,\%$. Ablations confirm that the
 heterogeneous graph, the MoE fusion layer and the fine-tuned backbone each
 contribute substantively to the output and the deep-ensemble UQ stack achieves
 post-calibration ECE of $0.031$ (hybrid) and $0.026$ ($\beta$-NLL).

\par To the best of our knowledge, this work introduces a novel framework that
unifies foundation-model feature extraction, heterogeneous graph attention and
decomposed uncertainty quantification for robust \textbf{\soc{}} estimation.
\end{abstract}

\section{Introduction}\label{sec:introduction}
\par \textbf{Soil organic carbon \soc{}} contains the largest terrestrial carbon
pool, storing approximately 1\,500\,Gt of carbon in the upper metre of soil,
exceeding the combined carbon stocks of the atmosphere and vegetation, stated by
\citep{totalcarbonandnitrogen}. This motivates the need for granular, accurate
\soc{} predictions for national greenhouse gas inventories, \textit{land use}
planning and \textit{carbon credit} markets.

\par Conventional digital soil mapping follows the \textit{\textbf{scorpan}}
framework \citep{mcbratney_digital_2003}, which trains \textit{tree-based
ensembles} on hand crafted \textit{environmental covariates} such as spectral
indices, terrain characteristics and climate variables
\citep{hengl_soilgrids250m_2017}. These pipelines suffer from the following
limitations: hand crafted features discard most of the high-dimensional spectral
and temporal information available in current satellite technologies and
standard regressors treat each sample in isolation, ignoring the fact that
nearby soils are correlated (spatial autocorrelation).

\par Recent developments in machine learning help us address these limitations.
\textit{Convolutional neural networks (CNNs)} learn image based representations
and patterns \citep{padarian_using_2019} from satellite sensors, but they use a
grid based representation, which is not able to accommodate the uneven and
irregular patterns that we get from soil surveys. This is where \textit{graph
neural networks (GNNs)} shine, representing each sample as a node in the graph
and associating samples by geographic location (proximity).
\citet{klemmer2023pegnn,zhao2023soc} propose \emph{graph neural networks for
geospatial regression}, which account for the sample spatial structure. More
recently, \emph{Earth-observation foundation models}, pretrained on large-scale
multi-sensor imagery have emerged as a useful way of extracting high level
information for remote sensing tasks. One such model is \textbf{TerraMind}
\citep{jakubik2025terramind}, an any-to-any generative multi-modal foundation
model pre-trained on nine satellite modalities. In addition, deep ensembles with
heteroscedastic likelihoods
\citep{lakshminarayanan2017ensembles,seitzer2022betanll} allow for calibrated,
decomposable uncertainty estimates rather than point predictions.

\par We propose \textbf{\sptgnn{}}, a \textit{multi-modal spatio-temporal
heterogeneous graph neural network} for \soc{} prediction, which combines
fine-tuned TerraMind satellite image embeddings, environmental covariates,
spatial and temporal encodings into a unified framework. Given these very
different streams of data, we use a cross-gated \textbf{Mixture-of-Experts
(MoE)} module to fuse the modalities, which then get fed into a heterogeneous
relational graph-attention network that propagates information across multiple
edge types capturing geographic, spectral and elevation similarity. Calibrated,
decomposed uncertainty is obtained through a deep ensemble with post-hoc
temperature scaling.

\paragraph{Contributions}
\par Our specific contributions are:
\begin{itemize}
    \item A \textbf{heterogeneous graph} construction, using three edge
    relations (geographic adjacency, NDVI spectral similarity and
    elevation), enhancing the GNN's capability to capture soil specific
    relationships. Previous literature \soc{} GNNs use a single spatial edge
    type, which in our experiments, dropping the NDVI and elevation relations costs
    ${\sim}0.39$ in $R^{2}$ on Africa set (Section \ref{ssec:ablations}),
    confirming that the three relations carry meaningful signal
    \item A \textbf{cross-gated Mixture-of-Experts fusion layer} that integrates
    per-region fine-tuned TerraMind embeddings with tabular, positional and
    temporal features, regularized by a load-balancing auxiliary loss. Replacing
    it with a concatenation MLP fusion costs ${\sim}0.29$ in $R^{2}$ on the same
    ablation benchmark
    \item An end-to-end \textbf{multi-task objective} that combines a target
    regression loss, a Moran's-$I$ auxiliary loss for spatial autocorrelation
    (after \citealt{klemmer2023pegnn}) and an \textit{optional} heteroscedastic
    Gaussian NLL with $\beta$-NLL (after \citealt{seitzer2022betanll}) for
    aleatoric uncertainty estimation
    \item A \textbf{deep ensemble UQ stack}, built on
    \citep{lakshminarayanan2017ensembles}, that decomposes total predictive
    uncertainty into aleatoric and epistemic uncertainty, with post-hoc
    temperature scaling for calibration (Section~\ref{ssec:uq-results})
\end{itemize}

\par Evaluation across \textbf{three regional datasets} from a diverse global
multi-source \soc{} corpus ($\sim$49\,k samples), with the best performing
Africa instance used as the focus of a detailed accuracy, ablation and
uncertainty analysis against a tabular gradient-boosting baseline. There, the
deep-ensemble \sptgnn{} reaches $R^{2}=0.762$ and MAPE\,$=22.9\,\%$ on the test
split and the best single-model checkpoint reaches a validation $R^{2}=0.864$
and MAPE\,$=13.7\,\%$.

The remainder of the paper is organized as follows, starting with
Section~\ref{sec:related-work} which positions \sptgnn{} in the literature on
\soc{} prediction, EO foundation models, geospatial graph learning, multi-modal
fusion techniques and uncertainty quantification. Section~\ref{sec:method}
details the \sptgnn{} architecture, the per-region ViT fine-tune, the
heterogeneous graph construction and the multi-task training objective.
Section~\ref{sec:dataset} breaks down the global \soc{} corpus and the remote
sensing data pipeline. Section~\ref{sec:experiments} reports accuracy results,
trained regional models, architecture ablations and uncertainty calibration on
our \textit{Africa} model. Section~\ref{sec:discussion} discusses limitations
and future directions.

\section{Related Work}\label{sec:related-work}
\par This section positions our proposed model within the landscape of
geospatial machine learning. We first trace the progression of \soc{} prediction
methods from classical covariates to graph-based architectures
(\ref{ssec:soc-prediction}), then review Earth observation foundation models as
feature extractors
(\ref{ssec:feature-extraction-via-earth-observation-foundation-models}), discuss
graph neural network architectures for spatial modeling
(\ref{ssec:graph-neural-networks-for-spatial-modeling}) and, at the end cover
techniques for multi-modal fusion and uncertainty quantification
(\ref{ssec:multimodal-fusion-and-uncertainty-quantification}).

\subsection{Soil Organic Carbon Prediction: From Covariates to
Graphs}\label{ssec:soc-prediction}
\par Mapping \soc{} on a regional and global scale sits at the core of carbon
accounting, land management policy and climate change mitigation, given that
soil carbon sequestration has the potential to offset 5–15\% of global fossil
fuel emissions \citep{lal_soil_2004}. The \textit{scorpan-SSPFe (soil spatial
prediction function with spatially autocorrelated errors)} framework
\citep{mcbratney_digital_2003} formalized the prediction of soil properties from
environmental covariates. Ensemble tree methods, specifically \textit{Random
Forests} and \textit{Gradient Boosting} became the dominant approach for large
scale applications such as \textbf{SoilGrids250m}
\citep{hengl_soilgrids250m_2017}.
\par Deep learning has since entered the field, with \textit{convolutional
neural network (CNN)} architectures, which improve prediction accuracy by taking
advantage of the spatial context from neighboring pixels
\citep{padarian_using_2019}. However, this network architecture requires regular
grids and cannot account for the unstructured geometry of sparse field collected
samples. In contrast, \textit{Graph neural networks (GNNs)} can model
relationships between irregularly placed samples. \citet{zhao2023soc} and
\citet{flores_graph_2026} apply \textbf{GNNs} to \soc{} prediction, proving that
graph-based approaches can capture spatial dependencies more effectively than
conventional methods, although both works rely on homogeneous graphs.

\subsection{Feature Extraction via Earth Observation Foundation
Models}\label{ssec:feature-extraction-via-earth-observation-foundation-models}

\par Extracting features from satellite sensor data, specifically images, was
traditionally constrained to task specific neural networks or spectral indices.
To overcome these limitations, self-supervised pre-training has produced several
Earth observation foundation models, including Prithvi-EO-2.0
\citep{szwarcman_prithvi-eo-20_2026} and AlphaEarth
\citep{brown_alphaearth_2025}, among which we adopt \textbf{TerraMind}
\citep{jakubik2025terramind}, an any-to-any generative multimodal foundation
model for EO. The model was pre-trained on nine modalities via dual-scale early
fusion. This fusion strategy produces two complementary representation types,
tokens that capture cross modal correlations and pixel level embeddings that
take into account fine-grained spatial detail. Together, these representations
allow TerraMind to generalize effectively across a wide range of downstream EO
tasks.
\par In this work, the frozen encoder is used to extract embeddings for each
given sample. However, these models treat each patch independently and do not
capture spatial dependencies between sample pairs. Combining their embeddings
with graph-based spatial reasoning remains unexplored for \soc{} estimation.

\subsection{Graph Neural Networks for Spatial
Modeling}\label{ssec:graph-neural-networks-for-spatial-modeling}

\par Modeling spatial relationships between geographic sites is one of the core
challenges in geospatial analysis, as conventional architectures generally
assume independent and identically distributed inputs, lacking the capacity to
encode pairwise relationships across arbitrary locations. \textbf{Graph neural
networks (GNNs)} address this by representing measurement sites as nodes and
their relationships as edges, with foundational operators such as \textbf{Graph
Convolutional Networks (GCN)} \citep{kipf_semi-supervised_2017} and
\textbf{Graph Attention Networks (GAT)} \citep{velickovic_graph_2018}
aggregating information across node neighbourhoods.

\par In the context of estimating soil properties, \citet{zhao2023soc} proved
the utility of this approach for the prediction of \soc{}, using the Positional
Encoder GNN (PE-GNN) framework \citep{klemmer2023pegnn}, solving spatial
coherence through a Moran’s~$I$ \citep{moran1950} regularization term. Their
results show that \textit{PE-SAGE} and \textit{PE-Transformer} outperform
\textit{PE-GCN} and \textit{PE-GAT}. However, the framework is limited to a
homogeneous graph with a single edge type and attention head and relies on hand
crafted covariates rather than learned representations.

\par To overcome the limitations of homogeneous graphs,
\citet{busbridge2019rgat} proposed the \textbf{Relational Graph Attention
Network} (\rgat{}), which assigns each edge type its own projection matrices and
attention vectors, so that structurally different relationships receive
independent attention weights. To date, such relational graph architectures have
not been explored for soil property estimation.

\subsection{Multi-modal Fusion and Uncertainty
Quantification}\label{ssec:multimodal-fusion-and-uncertainty-quantification}

\par Integrating multiple data sources effectively requires both preserving
their individual structure and enabling cross-modal interaction. Standard
\textit{early fusion} based on simple concatenation tends to ignore the
underlying structure, whereas late fusion restricts cross modal interactions and
limits the learning of joint representations. Sparse Mixture-of-Experts (\moe{})
layers \citep{shazeer2017moe} route individual samples to specialised experts
via a learned gating mechanism, combining heterogeneous inputs without a
proportional increase in computation. While effective in vision
\citep{riquelme2021vmoe} and language settings, their application to multimodal
remote sensing data remains limited.

\par Predictive models require both accurate outputs and \textbf{reliable
uncertainty estimates}. Deep ensembles \citep{lakshminarayanan2017ensembles}
provide \textbf{epistemic uncertainty}, which captures the model’s uncertainty
due to limited knowledge from training data, without requiring explicit Bayesian
posteriors. \textbf{Aleatoric uncertainty} accounts for the randomness in the
data and is usually modeled using a heteroscedastic Gaussian negative
log-likelihood. To stabilize the gradients in this standard formulation,
\citet{seitzer2022betanll} introduced $\beta$-NLL.

\par To the best of our knowledge, no existing framework jointly integrates EO
foundation model embeddings, heterogeneous graph attention, multi-modal fusion,
and decomposed uncertainty quantification for soil property estimation.

\section{Method}\label{sec:method}
\subsection{Problem Formulation}
\label{ssec:problem-formulation}

\par
\begin{figure}[htbp]
    \centering
    \includegraphics[width=1\textwidth]{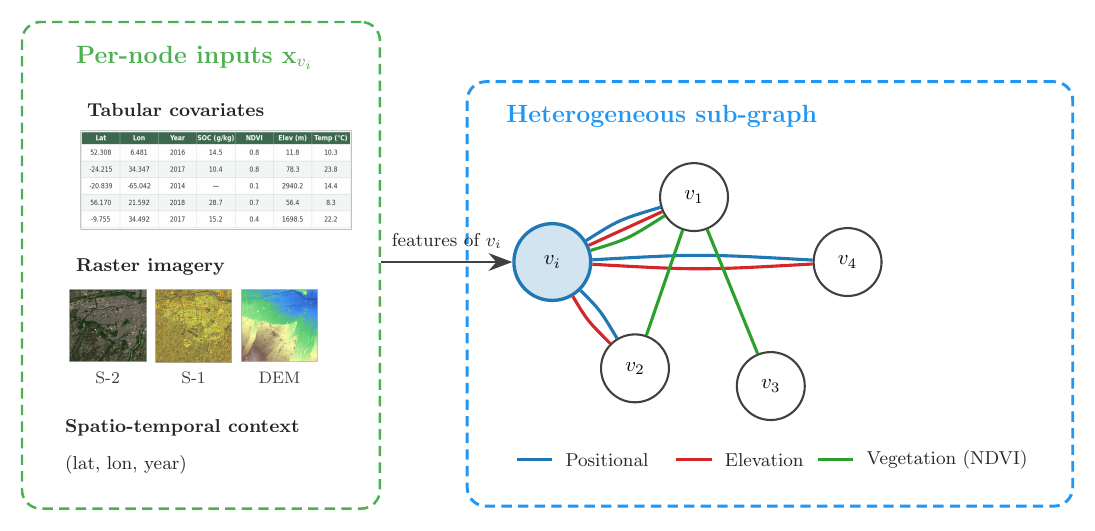}
    \caption{\textbf{Heterogeneous graph structure.} Nodes fuse multi-modal
    data, while relational edges \citep{busbridge2019rgat} capture geographic
    proximity, vegetation similarity and elevation similarity between
    samples.}
    \label{fig:graph_structure}
\end{figure}

 We structure the prediction of \soc{} as a graph based regression task,
outlined in Figure~\ref{fig:graph_structure}. We define the set $N$ of soil
samples $\{(\mathbf{x}_i, \mathbf{I}_i, \mathbf{p}_i, t_i, y_i)\}_{i=1}^{N}$. As
the figure depicts, each sample encapsulates multi-modal inputs: $\mathbf{x}_i
\in \mathbb{R}^{d}$ are tabular covariates, $\mathbf{I}_i$ represents satellite
imagery (Sentinel-2 optical, Sentinel-1 SAR and a digital elevation model DEM),
$\mathbf{p}_i = (\text{lon}, \text{lat})$  for the geographic coordinates, $t_i$
is the integer corresponding to the sampling year and $y_i \in \mathbb{R}$ is
the corresponding ground truth \soc{} measurement.

\par We represent these samples as nodes within a heterogeneous graph
$\mathcal{G} = (\mathcal{V},\, \mathcal{E}_s,\, \mathcal{E}_v,\,
\mathcal{E}_e)$, with three edge types that encode: relative position
($\mathcal{E}_s$), vegetation similarity ($\mathcal{E}_v$) and relative
elevation ($\mathcal{E}_e$). Each of these edge types are expanded in
Section~\ref{ssec:graph-construction}.

\par We train the model on a single graph whose topology is shared across data
splits, masking the \textbf{validation} and \textbf{test} targets. By doing
this, we allow the message passing to propagate through unlabeled nodes. At
inference time, predictions for unseen locations are obtained by constructing a
local subgraph from the target point's nearest neighbors in the training graph
(Section~\ref{ssec:graph-construction}).

\par Our model produces three outputs for a given node, a point
\textbf{prediction} $\hat{y}_i$, a \textbf{spatial-autocorrelation auxiliary}
$\hat{\iota}_i$ and optionally a \textbf{log-variance estimate} $s_i = \log
\sigma_i^2$ for predictive uncertainty.

\par This formulation builds on the observation that \soc{} varies with
geographic location, vegetation cover and elevation. Temporal context is
captured through encodings of the sampling year.

\subsection{Model Overview}
\label{ssec:model-overview}

\par
\begin{figure}[htbp]
    \centering
    \includegraphics[width=1\textwidth]{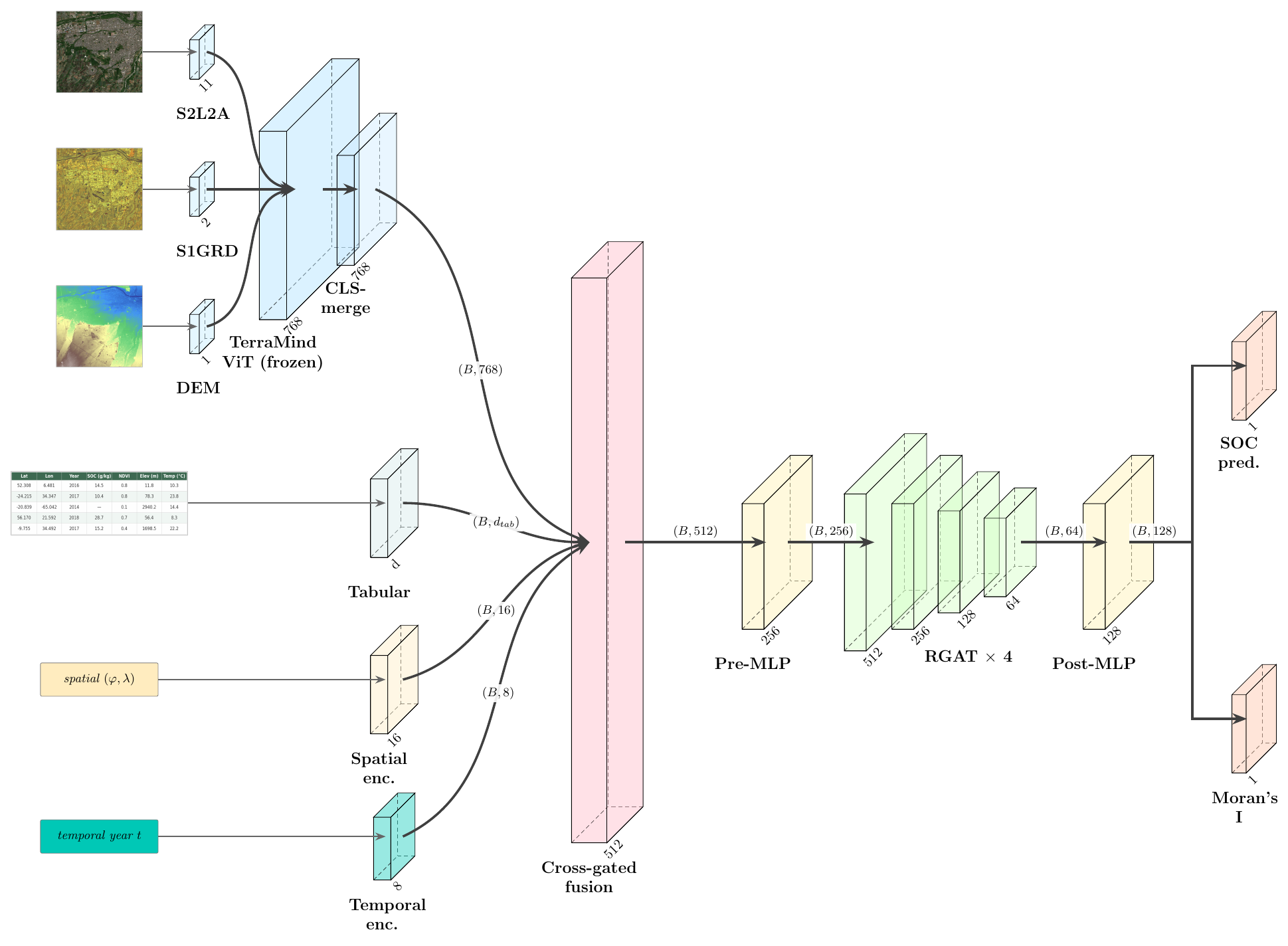}
    \caption{\textbf{SpTGNN architecture}. Multi-modal inputs (S2L2A,
    S1GRD, DEM, tabular, $(\varphi,\lambda)$, $t$) are embedded by
    modality-specific encoders, a frozen, per-region fine-tuned
    TerraMind ViT with CLS merge for imagery and MLPs for the rest,
    fused by a cross-gated 8-expert MoE layer and propagated
    through a four-layer heterogeneous RGAT stack over three edge
    types (geographic, NDVI, elevation). Two dense heads output the \soc{}
    prediction and a training only Moran's-$I$ auxiliary.}
    \label{fig:architecture}
\end{figure}

 The design principle that sits at the core of \sptgnn{} is the separation
of \textit{what a sample looks like} from \textit{how it relates to its
neighbours}. As a result, the encoding of each node feature and the relational
graph are handled sequentially. The forward pass consists of four stages
(Figure~\ref{fig:architecture}; pseudo-code in Algorithm~\ref{alg:forward}):

\begin{enumerate}
    \item \textbf{Multi-modal Encoding
    (Section~\ref{ssec:multi-modal-encoding}):} Each node's input is processed
    independently. Satellite imagery is passed through a pretrained vision
    transformer (TerraMind), geographic coordinates and sampling year are mapped
    to continuous embeddings using \textit{sinusoidal encoders} and tabular
    covariates are ingested after pre-processing.
    \item \textbf{Sparse Fusion (Section~\ref{ssec:moe-fusion}):} A sparse
    mixture-of-experts (MoE) layer fuses the input streams into a single node
    representation, routing each sample to the most relevant expert network.
    \item \textbf{Relational Reasoning (Sections~\ref{ssec:graph-construction}
    and \ref{ssec:relational-gat}):} A stack of relational graph attention
    (RGAT) layers propagates information from the heterogeneous graph,
    performing message passing over spatial, vegetation and elevation edges.
    \item \textbf{Multi-task Prediction (Sections~\ref{ssec:learning-objective}
    and \ref{ssec:uncertainty}):} Three linear heads map the final node
    embeddings to the main \textbf{\soc{} prediction}, a \textbf{spatial
    autocorrelation auxiliary} term and an \textit{optional}
    \textbf{log-variance} for uncertainty estimation.
\end{enumerate}

\begin{algorithm}[t]
  \caption{\sptgnn{} Forward Pass}\label{alg:forward}
  \KwIn{Heterogeneous graph $\mathcal{G}=(\mathcal{V},\{\mathcal{E}_r\}_{r\in\mathcal{R}})$;
    per-node features $\{\mathbf{x}_i, \mathbf{I}_i, \mathbf{p}_i, t_i\}_{i\in\mathcal{V}}$}
  \KwOut{Predictions $\{\hat{y}_i,\, \hat{\iota}_i,\, \sigma_i^2\}_{i\in\mathcal{V}}$}

  \BlankLine
  \Comment{Per-node encoding}
  \For{each node $i \in \mathcal{V}$}{
    $\mathbf{h}_i^{\text{vit}} \gets \mathrm{ViTEncode}(\mathbf{I}_i)$
      \tcp*{frozen TerraMind + CLS-merge}
    $\mathbf{e}_i^{\text{pos}} \gets \mathrm{PosEncode}(\mathbf{p}_i)$
      \tcp*{Eq.~\eqref{eq:pos-encoding}}
    $\mathbf{e}_i^{\text{time}} \gets \mathrm{TimeEncode}(t_i)$
      \tcp*{Eq.~\eqref{eq:time-encoding}}
    $\mathbf{z}_i \gets
      [\mathbf{h}_i^{\text{vit}};\,\mathbf{x}_i;\,\mathbf{e}_i^{\text{pos}};\,\mathbf{e}_i^{\text{time}}]$\;
  }

  \BlankLine
  \Comment{Sparse Mixture-of-Experts fusion}
  \For{each node $i \in \mathcal{V}$}{
    $\ell_i \gets W_g \,\mathrm{LN}(\mathbf{z}_i) + \mathbf{b}_g$
      \tcp*{gating logits}
    $\mathcal{T}_i \gets \mathrm{Top}_k(\ell_i)$\;
    $g_{i,e} \gets \mathrm{softmax}(\ell_{i,\mathcal{T}_i})_e$ for $e\!\in\!\mathcal{T}_i$\;
    $\mathbf{u}_i \gets \sum_{e\in\mathcal{T}_i} g_{i,e}\, f_e(\mathbf{z}_i)$\;
    $\mathbf{h}_i^{(0)} \gets \mathrm{PreMLP}(\mathbf{u}_i)$\;
  }

  \BlankLine
  \Comment{Relational GAT stack}
  \For{$\ell = 1$ \KwTo $L$}{
    \For{each node $i \in \mathcal{V}$}{
      $\mathbf{h}_i' \gets
        \mathrm{RGAT}\!\bigl(\mathbf{h}_i^{(\ell-1)},
          \{\mathbf{h}_j^{(\ell-1)} : j\in\mathcal{N}_r(i),\,r\in\mathcal{R}\}\bigr)$\;
      $\mathbf{h}_i^{(\ell)} \gets
        \mathrm{LN}\bigl(\mathrm{Skip}(\mathbf{h}_i^{(\ell-1)}) + \mathbf{h}_i'\bigr)$\;
    }
  }

  \BlankLine
  \Comment{Post-processing and prediction heads}
  \For{each node $i \in \mathcal{V}$}{
    $\mathbf{h}_i^{\star} \gets \mathrm{PostMLP}(\mathbf{h}_i^{(L)})$\;
    $\hat{y}_i \gets \mathrm{Head}_y(\mathbf{h}_i^{\star})$\;
    $\hat{\iota}_i \gets \mathrm{Head}_\iota(\mathbf{h}_i^{\star})$\;
    $s_i \gets s_{\min} + (s_{\max}-s_{\min})\cdot
      \tfrac{1}{2}\bigl(1+\tanh\bigl(\mathrm{Head}_\sigma(\mathbf{h}_i^{\star})\bigr)\bigr)$
      \tcp*{if UQ enabled}
  }

  \Return $\{\hat{y}_i,\; \hat{\iota}_i,\; \exp(s_i)\}_{i\in\mathcal{V}}$
  \end{algorithm}

\subsection{Multi-Modal Feature Encoding}
\label{ssec:multi-modal-encoding}

\par Each node's information is encoded by separate branches that operate
independently before fusion:

\paragraph{Satellite imagery}
As mentioned previously, we use TerraMind foundation model by
\citep{jakubik2025terramind}, which is pre-trained on multi-modal Earth
observation data.

Before integrating TerraMind into our pipeline, we first fine-tune the
foundation model backbone on our \soc{} dataset as a regression task in order to
adapt its representations for our domain and target. Then we use the resulting
frozen weights as a \textit{feature extractor}, part of our \sptgnn{} model. For
each node, Sentinel-2 (optical), Sentinel-1 (SAR) and DEM patches centered on
the sample location are encoded independently by the frozen backbone. This
generates a patch embedding sequence for each modality. We merge these sequences
with a per-region element-wise reduction across the modalities (mean for the
Africa and Europe instances, maximum for the Global instance, see
Table~\ref{tab:hyperparams}). We then apply \textit{max-pooling} over the patch
dimension to produce a single vector $\mathbf{h}_i^{\text{vit}} \in
\mathbb{R}^{d_{\text{vit}}}$.

\paragraph{Geographic position}

Using raw \textit{lat/lon} coordinates fails to represent true spatial
relationships on a \textit{spherical} shape and lack a consistent scale (small
changes in latitude or longitude can correspond to large physical distances).

We encode each coordinate pair via multi-scale sinusoidal features, followed by
a two layer perceptron:
\begin{equation}\label{eq:pos-encoding}
    \mathbf{e}_i^{\text{pos}}
      = \mathrm{MLP}\!\Bigl(
          \bigl[\sin(\mathbf{p}_i / \sigma_k),\;
                \cos(\mathbf{p}_i / \sigma_k)\bigr]_{k=1}^{K}\Bigr),
\end{equation}
where $\{\sigma_k\}_{k=1}^{K}$ are logarithmic scales $K$ that span
$[\sigma_{\min},\, \sigma_{\max}]$.

This allows the network to interpret exact geographic locations across both
narrow and wide scales. We adopt the \textit{deterministic sinusoidal encoding}
strategy of \citet{mildenhall_nerf_2020}, which \citet{tancik_fourier_2020}
proved to be necessary to overcome the limitation that standard neural networks
struggle to learn complex spatial patterns directly from simple lat/lon
coordinates.

\paragraph{Temporal context}
The sampling year $t_i$ is normalized to $[0,1]$ and encoded with
\textit{Fourier features} at log-spaced frequencies:
\begin{equation}\label{eq:time-encoding}
    \mathbf{e}_i^{\text{time}}
      = \mathrm{MLP}\!\Bigl(
          \bigl[\sin(\bar{t}_i \cdot f_j),\;
                \cos(\bar{t}_i \cdot f_j)\bigr]_{j=1}^{J}
        \Bigr),
\end{equation}
where $\bar{t}_i = (t_i - t_{\min}) / (t_{\max} - t_{\min})$.

This allows the model to track changes over time without dividing the years into
separate time bins.

\paragraph{Tabular covariates}
The tabular data $\mathbf{x}_i$ requires no learned encoding and is passed
directly to the fusion stage.

\par The next subsection explains how we merge these four different data streams
into a unified representation.

\subsection{Sparse Mixture-of-Experts Fusion}
\label{ssec:moe-fusion}

\par Instead of concatenating the four input streams, which would treat each
feature equally, we use a \textit{sparse mixture-of-experts (MoE)} layer that
adapts to the input.

\par
\begin{figure}
    \centering
    \includegraphics[width=1\textwidth]{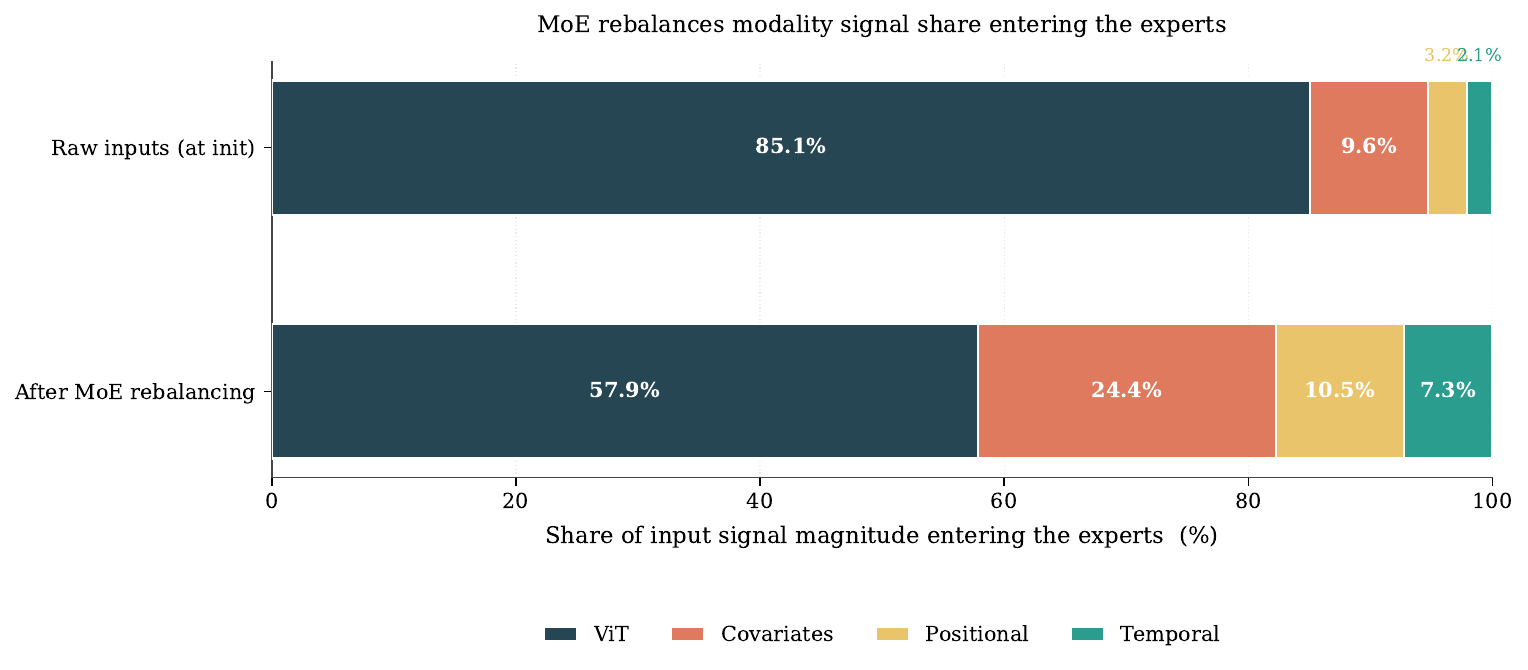}
    \caption{Share of input signal magnitude entering the experts, before and
    after MoE rebalancing: ViT dominates the raw inputs (85\%), while
    rebalancing redistributes signal for covariates, positional and temporal
    modalities}
    \label{fig:feature-magnitudes}
\end{figure}

 Because the inputs have different scales
(Figure~\ref{fig:feature-magnitudes}), we first apply layer normalization for
each input, making sure that no modality dominates the routing. The normalized
features are then concatenated and passed to a gating network. Following
\citep{shazeer2017moe}, the neural network uses \textit{dynamic routing} to
select only the top-$k$ experts for each node, setting the rest of the gates
zero.

\par In essence, each expert is a \textit{dedicated network} that combines the
different data streams using learned gates. As a result, the final node
embedding is the \textit{gate weighted} sum of the expert's outputs. A possible
issue that may arise during the process is \textit{expert collapse}, where the
network ends up relying on just one expert and ignores the rest. We include two
regularizers in the main training loss (Section~\ref{ssec:learning-objective}),
a \textit{load balancing term} to enforce even expert usage for each batch and
an \textit{entropy penalty} to diversify selections.

While MoE models are more common in language processing, applying them to
multi-modal geospatial data is, to our knowledge, a novel approach. We compare
this with a simple concatenation fusion method in Section~\ref{ssec:ablations}.

\subsection{Heterogeneous Graph Construction}
\label{ssec:graph-construction}

\par After having the node embeddings fused, we define the topology of the
graph. Previous works that use GNN models for \soc{} prediction,
\citep{zhao2023soc} implement homogeneous spatial graphs, with one edge type
accounting for sample proximity. Building on this work, we encode environmental
factors directly into the graph structure as distinct edge types. This results
in a heterogeneous graph that models three types of associations (Table
\ref{tab:edge-relations}).

\begin{table}[t]
\centering
    \caption{Edge relations in the heterogeneous graph.
  Neighbor counts $k$ and scale parameters $\tau$ are reported in
  Table~\ref{tab:hyperparams}}
\label{tab:edge-relations}
\small
\begin{tabular}{@{}llll@{}}
  \toprule
  Relation & Neighborhood & Edge weight & Rationale \\
  \midrule
  Spatial $\mathcal{E}_s$
    & $k$-NN in $(\text{lon},\text{lat})$
    & Inverse distance
    & Geographic proximity \\
  Vegetation $\mathcal{E}_v$
    & $k$-NN in NDVI space
    & $\exp(-|\Delta\text{NDVI}|/\tau_v)$
    & Similar land cover \\
  Elevation $\mathcal{E}_e$
    & Spatial edges, re-weighted
    & $\exp(-|\nabla z|/\tau_t)$
    & Elevation \\
  \bottomrule
\end{tabular}
\end{table}

\par Spatial edges ($\mathcal{E}_s$) are set to connect geographically close
\soc{} samples, following Tobler's first law of geography
\citep{miller2004tobler}, stating that \textit{"everything is related to
everything else, but near things are more related than distant things"}.
Vegetation edges ($\mathcal{E}_v$) link samples with similar vegetation indices
(NDVI), grouping samples by land cover even if they are physically far apart
from each other. Elevation edges ($\mathcal{E}_e$) share the spatial structure
but reduce connection weights when there are considerable elevation changes.

\par Using a single graph would blend these distinct types of similarity. By
separating them into typed relations, the downstream relational attention
mechanism \citep{busbridge2019rgat} can evaluate geographic, ecological and
topographic influences independently (we evaluate this choice in
Section~\ref{sec:experiments}).

\subsection{Relational Graph Attention Network}
\label{ssec:relational-gat}

\par
\begin{figure}[t]
\centering
\includegraphics[width=\textwidth]{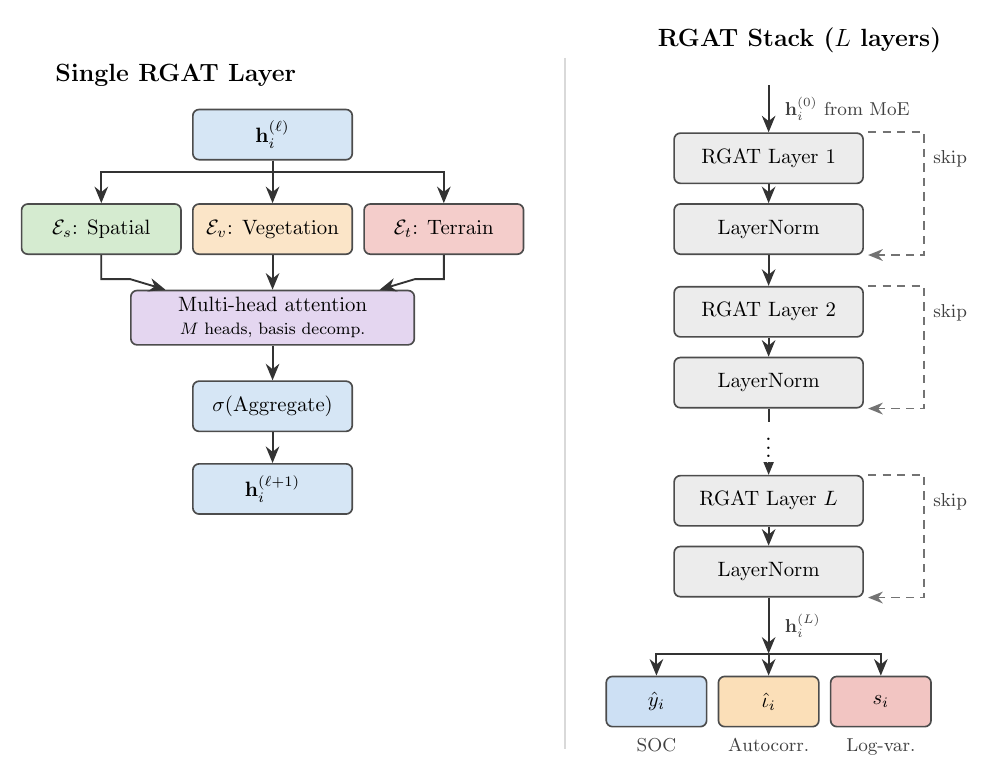}
\caption{%
  \textbf{Left:} A single RGAT layer computes relation aware multi-head
  attention over three edge types (spatial, vegetation, elevation), then
  aggregates weighted messages through a nonlinearity.
  \textbf{Right:} The full stack repeats $L$ such layers, each with a
  residual skip connection and layer normalization.  Three linear heads
  map the final embedding $\mathbf{h}_i^{(L)}$ to the SOC prediction,
  a spatial-autocorrelation auxiliary and an optional log-variance.
}
\label{fig:rgat-stack}
\end{figure}

 The fused node embeddings are processed by a stack of relational graph
attention (RGAT) layers \citep{busbridge2019rgat}. Through these layers, the
model can route information through the graph's nodes based on the different
connection types (Figure~\ref{fig:rgat-stack}).

\paragraph{Message passing}
First, we combine all edges into a set, labeled by relation type. Then in each
layer, nodes aggregate information from their neighbors using \textit{multi-head
attention}. The attention mechanism is relation specific, meaning that a
neighbor's relevance depends on both the node features and the type of edge
connecting them.  As the number of parameters increases linearly with the edge
relation types and attention heads, we use \textbf{basis decomposition}
\citep{busbridge2019rgat}.

\paragraph{Stack design and solving oversmoothing}
The model processes the graph through $L$ sequential RGAT layers (Figure
\ref{fig:rgat-stack}). The forward pass for each layer consists of a
\textbf{relational graph convolution}, a \textbf{nonlinear activation}, a
\textbf{residual addition} and \textbf{layer normalisation}. GNNs are prone to
oversmoothing, as stated by \citep{li_deeper_2018}, where nodes gradually lose
their distinct features. To overcome this, we implement the following
mechanisms:
\begin{itemize}
    \item residual skip connections with learned linear projections
    \item layer normalization
\end{itemize}

\paragraph{Prediction heads}
Three linear projections map the final node embedding to outputs:
\begin{itemize}
    \item $\hat{y}_i$ for \soc{} regression
    \item $\hat{\iota}_i$ for a spatial autocorrelation auxiliary
    \item $s_i$ for log-variance when uncertainty estimation is enabled
    (Section~\ref{ssec:uncertainty}).
\end{itemize}

The autocorrelation head acts as a \textbf{structural regularizer}, requiring
the network to predict a local Moran index \citep{zhao2023soc} alongside the
\soc{} value, by doing this, we push the learned representation to encode
spatial context, rather than relying solely on point features.

\subsection{Multi-Task Learning Objective}
\label{ssec:learning-objective}

\par The model is trained by minimizing a \textbf{combined loss}:

\begin{equation}\label{eq:loss}
\mathcal{L}
  = \mathcal{L}_{\text{main}}
  + \lambda_a \,\mathcal{L}_{\text{autocorr}}
  + \mathcal{L}_{\text{MoE}}.
\end{equation}

where the individual terms are defined as follows:
\begin{itemize}
  \item $\mathcal{L}_{\text{main}}$ is \textit{Huber loss}
  \citep{huber_robust_1992} between the predicted and observed \soc{}. We use it
  because of its robustness against outliers.
  \item $\mathcal{L}_{\text{autocorr}}$ applies the same Huber loss to predict
  local Moran's~$I$ values for each node,  weighted by $\lambda_a$, pushing the
  network to learn broad spatial relationships.
  \item $\mathcal{L}_{\text{MoE}}$ combines the load-balancing and entropy
  penalties from the MoE layer (Section~\ref{ssec:moe-fusion}). With
  $\bar{u}_e$ the mean gate weight of expert $e$ over the batch and $E$ the
  number of experts, the load-balancing term is
  $\mathcal{L}_{\text{LB}} = \frac{1}{E}\sum_{e}\bigl(\bar{u}_e - 1/E\bigr)^2$
  and the entropy penalty is
  $\mathcal{L}_{\text{ent}} = \log E + \sum_{e}\bar{u}_e \log \bar{u}_e$, i.e.,
  the entropy gap to uniform usage. We set
  $\mathcal{L}_{\text{MoE}} = w_{\text{LB}}\,\mathcal{L}_{\text{LB}} +
  \mathcal{L}_{\text{ent}}$, where $w_{\text{LB}}$ is the tuned per-region
  load-balance weight reported in Table~\ref{tab:hyperparams} and the entropy
  penalty enters with unit weight.
\end{itemize}
Standard weight regularization is done through the optimiser's decoupled weight
decay.

\subsection{Predictive Uncertainty Estimation}
\label{ssec:uncertainty}

\par Along with point predictions, the model can also estimate uncertainty for a
given output, as depicted on the right side of Figure \ref{fig:rgat-stack}.
\textbf{Aleatoric uncertainty} is captured by the heteroscedastic variance head
introduced in Section \ref{ssec:relational-gat}, while \textbf{epistemic
uncertainty} is the output of a small deep ensemble.

\par When uncertainty estimation is enabled, $\mathcal{L}_{\text{main}}$ in Eq.
\eqref{eq:loss} is augmented with (or replaced by) a \textit{Gaussian negative
log-likelihood}:

\begin{equation}\label{eq:nll}
\mathcal{L}_{\text{NLL}}
  = \frac{1}{2N} \sum_{i=1}^{N}
    \bigl[
      e^{-s_i}(y_i - \hat{y}_i)^2 + s_i
    \bigr].
\end{equation}

\par \citet{seitzer2022betanll} shows that Eq.~\eqref{eq:nll}'s mean prediction
gradient is scaled by $\sigma_i^{-2}$, meaning that samples with high variance
contribute negligibly to learning $\hat{y}_i$.
We adopt their $\beta$-NLL re-weighting:
\begin{equation}\label{eq:beta-nll}
  \mathcal{L}_{\beta\text{-NLL}}
    = \frac{1}{2N} \sum_{i=1}^{N}
      \mathrm{sg}\!\bigl[\sigma_i^{2\beta}\bigr]\,
      \bigl[
        e^{-s_i}\,(y_i - \hat{y}_i)^2 \;+\; s_i
      \bigr],
\end{equation}
where $\mathrm{sg}[\,\cdot\,]$ denotes the stop-gradient operator (the bracketed
term is treated as a constant during backpropagation, contributing no gradient).

\par We train an ensemble of multiple models following the same architecture,
but with different seeds, impacting the weight initialization and batch order.
The output predictions are aggregated via the law of total variance from
\citep{lakshminarayanan2017ensembles}:
\begin{align}\label{eq:ensemble}
  \hat{y}_i &\;=\; \frac{1}{M}\sum_{m=1}^{M} \hat{y}_i^{(m)},\\[2pt]
  \sigma_{\text{aleatoric}}^{2}(i)
            &\;=\; \frac{1}{M}\sum_{m=1}^{M} e^{s_i^{(m)}},\nonumber\\[2pt]
  \sigma_{\text{epistemic}}^{2}(i)
            &\;=\; \frac{1}{M-1}\sum_{m=1}^{M}
                  \bigl(\hat{y}_i^{(m)} - \hat{y}_i\bigr)^{2},\nonumber\\[2pt]
  \sigma_{\text{total}}^{2}(i)
            &\;=\; \sigma_{\text{aleatoric}}^{2}(i)
                 + \sigma_{\text{epistemic}}^{2}(i).\nonumber
\end{align}

Aleatoric variance is drawn from the mean variance of the predictions per
ensemble member. Epistemic variance is the \textit{Bessel-corrected} variance of
the ensemble predictions, capturing the model disagreement, which should grow in
sparse data regions.

\par The total predictive variance is calibrated with a single learned scalar $T
> 0$ (Eq. \ref{eq:temp}), which is fit on the \textit{validation} split,
minimising Eq.~\eqref{eq:nll} with the calibrated variance.
\begin{equation}\label{eq:temp}
  \tilde{\sigma}_i \;=\; T\,\sigma_{\text{total}}(i),
\end{equation}

\par Section~\ref{sec:experiments} reports PICP at $\alpha = 95\%$, ECE
as the mean absolute coverage gap across $20$ confidence levels,
CRPS under a Gaussian predictive and the epistemic fraction
$\sigma_{\text{epistemic}}^{2}/\sigma_{\text{total}}^{2}$.

\section{Dataset}\label{sec:dataset}
\subsection{Soil Organic Carbon Data Sources}
\label{ssec:soil-organic-carbon-data-sources}

\par Many existing GNN-based \soc{} studies, such as the ones mentioned
previously \citep{zhao2023soc,flores_graph_2026} focus on specific datasets like
LUCAS \citep{orgiazzi2018lucas}, which limits the geographic context and sample
diversity.

\par To build a more diverse and comprehensive model, we assemble a global
dataset, aggregating \soc{} measurements from multiple, well spread sources:
AfSIS (2017), LUCAS (2018) \citep{orgiazzi2018lucas}, the Chilean SOCDB,
iSDAsoil, MangrovesDB, SoDaH and SOCPDB. With the exception of our independently
surveyed field samples, all sets come from the unified \textbf{Open Compendium
of Soil Datasets} \citep{hengl_2025_soildb}. Because these datasets provide
broad geographic coverage, different sampling protocols and measuring depths, we
apply a harmonization process, following the steps:

\begin{itemize}
    \item Convert all \soc{} values to \textit{g/kg}
    \item Retain only topsoil measurements, from 0 to 30 cm
    \item Deduplicate overlapping samples
    \item Remove records with missing coordinates, implausible \soc{} values or
    incomplete metadata
    \item Reproject all locations to \textbf{WGS 84}
\end{itemize}

\par
\begin{figure}[t]
    \centering
    \begin{subfigure}[t]{0.58\textwidth}
    \centering
    \includegraphics[width=\textwidth]{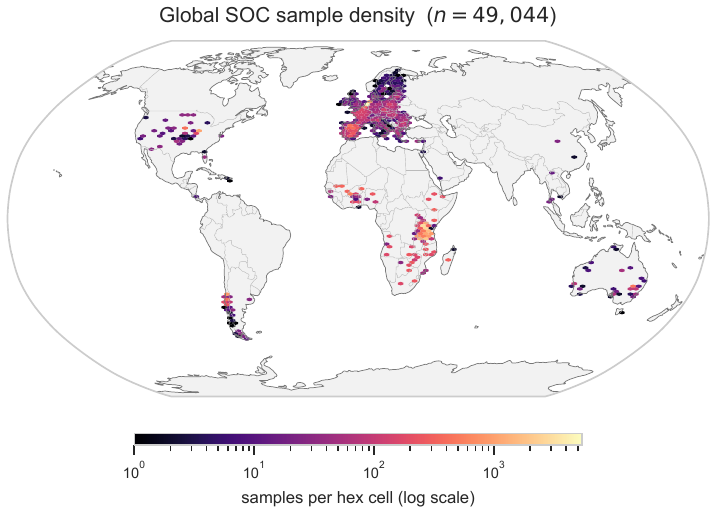}
    \caption{Global sample density (hex bins, log scale).}
    \label{fig:density-global}
    \end{subfigure}
    \hfill
    \begin{subfigure}[t]{0.38\textwidth}
    \centering
    \includegraphics[width=\textwidth]{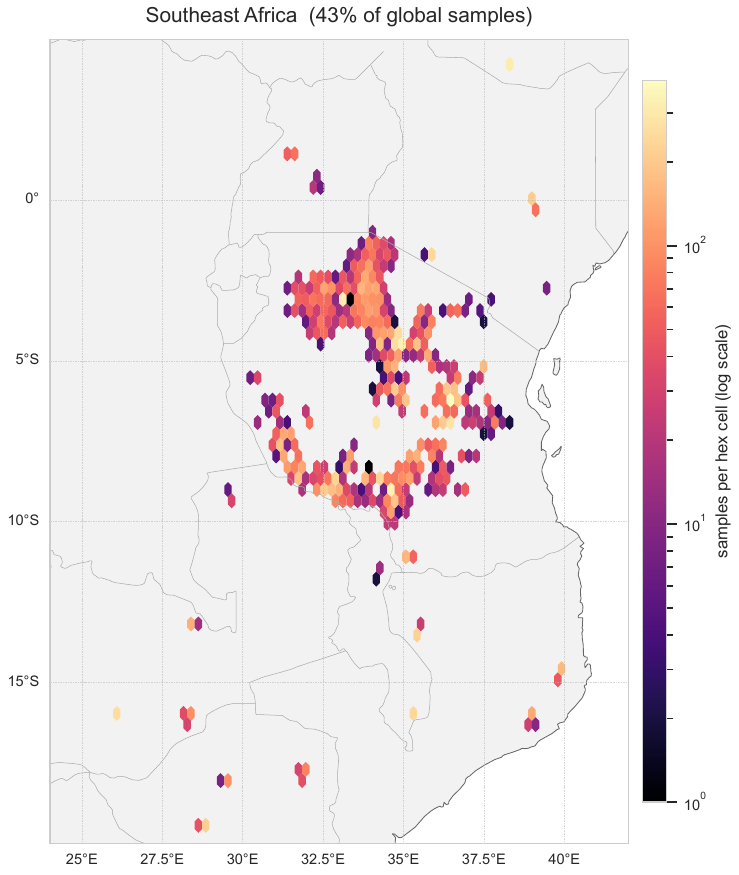}
    \caption{Southeast Africa inset (43\% of samples).}
    \label{fig:density-zoom}
    \end{subfigure}
    \caption{Spatial distribution of \soc{} samples. Density is shown as samples
    per hexagonal cell on a logarithmic colour scale.}
    \label{fig:sample-density}
\end{figure}

 After the harmonization process, the final dataset contains $N = 49{,}044$
soil samples spread across six continents. Figure~\ref{fig:sample-density} shows
the spatial distribution of these sample, with the densest being \textbf{Europe
(LUCAS)} and \textbf{Southeast Africa (AfSIS)}, with sparse representation in
South America, Australia and Southeast Asia.

\subsection{Remote Sensing Inputs}
\label{ssec:remote-sensing-inputs}

For each sample location, we extract localized satellite image patches from
three remote sensing modalities, temporally matched to the corresponding \soc{}
sampling year.

\paragraph{Sentinel-2 L2A (optical)}
We use \textbf{Level-2A Bottom-Of-Atmosphere (BOA)} multispectral imagery from
the harmonized Sentinel-2 Earth Engine collection. It consists of 11 spectral
bands (Blue, Green, Red, three vegetation red-edge, narrow and broad NIR, water
vapor and two SWIR) at 10-20\,m resolution. For each sample we build an annual
\textbf{median} from all acquisitions in the sampling year with cloud cover
below 25\%.

\paragraph{Sentinel-1 GRD (SAR)}
\textbf{Ground Range Detected (GRD)} images in VV and VH polarisations
(\texttt{COPERNICUS/S1\_GRD}) capture weather independent structural information
at 10\,m resolution. We use the standard pre-processing pipeline of Earth
Engine's GRD collection which consists of the steps: \textit{radiometric
calibration}, \textit{thermal noise removal} and \textit{range-Doppler terrain
correction}. Just as in the previous paragraph, for each sample, we compute an
annual median, which reduces variability.

\paragraph{Topography (DEM)}
Elevation data comes from the \textbf{Copernicus GLO-30 Digital Elevation Model}
(\texttt{COPERNICUS/DEM/GLO30}) at a 30\,m spatial resolution. On top of
elevation, we calculate local slope, which is used to compute the elevation
edge weights within the heterogeneous graph
(Section~\ref{ssec:graph-construction}).

\paragraph{Patch extraction}
For every \soc{} sample, we extract a square patch ($224 \times 224$ pixels)
centered around the target for all modalities. As this is a costly operation in
terms of time, we avoid processing bottlenecks during training by pre-computing
all patches and cache to tensors.

\begin{figure}[t]
  \centering
  \includegraphics[width=\textwidth]{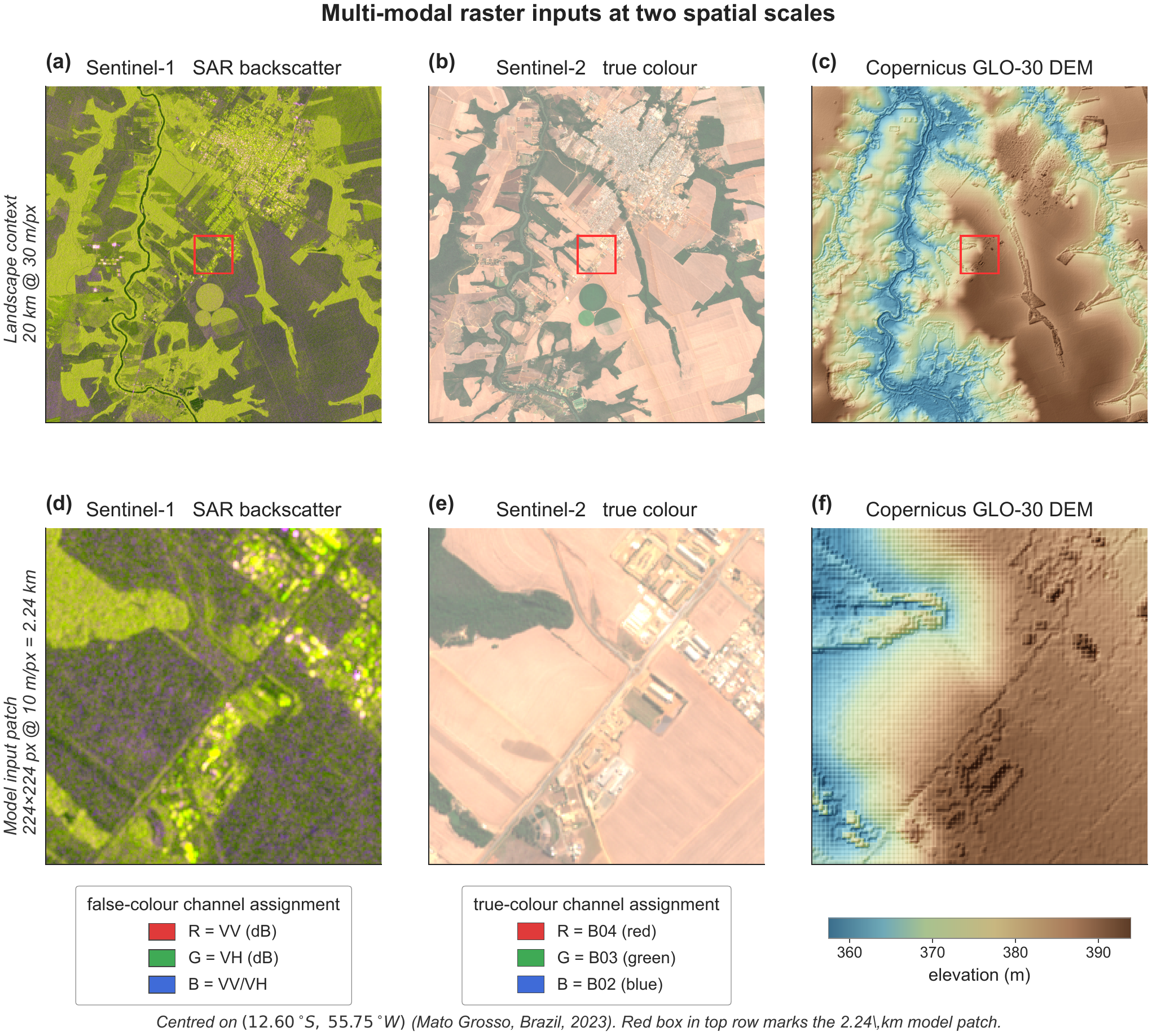}
  \caption{Multi-modal raster inputs at two spatial scales.
    \textbf{Top row (a--c):} landscape context at 30\,m/px over a
    20\,km region.
    \textbf{Bottom row (d--f):} the $224 \times 224$ pixel model input
    patch at 10\,m/px (2.24\,km per side), following ViT's
    encoder input format (Section~\ref{ssec:multi-modal-encoding}).
    The red box in the top row marks the patch extent.
    Location: Mato Grosso, Brazil.}
  \label{fig:modality-dual-scale}
\end{figure}

Figure~\ref{fig:modality-dual-scale} illustrates the three modalities at both
landscape scale (top row, 20\,km at 30\,m/px) and at the model input scale
(bottom row).

\subsection{Tabular Covariates}
\label{ssec:tabular-covariates}
\par Another input type that's received by each node, is \textbf{tabular
covariates}, which have many different un-encoded tabular features
(Section~\ref{ssec:multi-modal-encoding}) and can be split into eight thematic
groups as per Table~\ref{tab:covariate-groups}.

\par Tabular covariates with missing values are filled using
$k$-nearest-neighbour interpolation \citep{cover1967nearest} with $k = 5$. All
features go through a standardization process to zero mean and unit variance.
Both the imputer and scaler are fitted on training samples \textbf{only}, to
prevent data leakage.

\begin{table}[t]
  \centering
  \caption{Tabular covariate groups.  Percentile features (p10, p50, p90) are
  annual summaries derived from Sentinel-2 time series via Google Earth Engine.}
  \label{tab:covariate-groups}
  \small
  \begin{tabular}{@{}llr@{}}
    \toprule
    Group & Examples & Features \\
    \midrule
    Spectral statistics
      & Blue/Green/Red/NIR/SWIR p10, p50, p90; NDVI p10, p50, p90
      & 21 \\
    Phenology
      & Start/end/peak of season, EVI amplitude, cycle count
      & 11 \\
    Vegetation structure
      & LAI mean/std, fAPAR mean/std
      & 4 \\
    Topography
      & Elevation, slope, aspect, TPI, topographic diversity, CHI
      & 6 \\
    Precipitation climate
      & Mean annual, seasonality, driest/wettest/warmest month/quarter
      & 7 \\
    Temperature climate
      & Mean annual, diurnal/annual range, seasonality, quarterly means
      & 8 \\
    Land characteristics
      & Land use, parent material, agro-ecological zone
      & 3 \\
    Other
      & Fire frequency, bare surface frequency, NPP
      & 3 \\
    \midrule
    \textbf{Total} & & \textbf{63} \\
    \bottomrule
  \end{tabular}
\end{table}

\subsection{Dataset Statistics}
\label{ssec:dataset-statistics}

\par Following the harmonization of the global dataset, we end up with a diverse
range of soil conditions, as Table~\ref{tab:dataset-stats} reveals a summary of
the final set.

\par
\begin{figure}[t]
  \centering
  \begin{subfigure}[t]{0.48\textwidth}
    \centering
    \includegraphics[width=\textwidth]{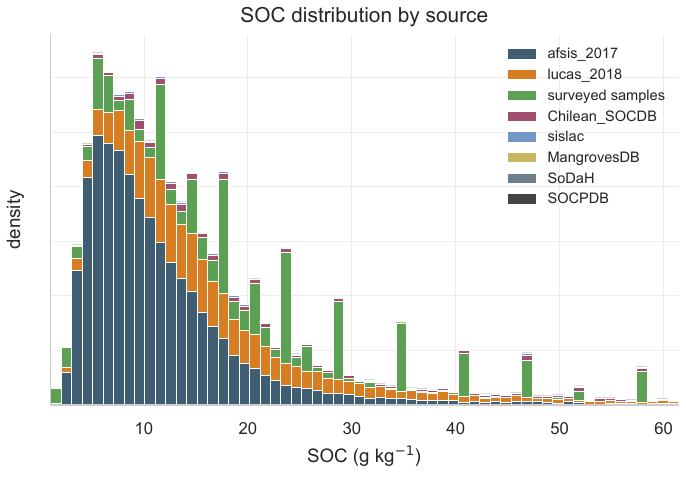}
    \caption{\soc{} distribution coloured by source database.
      AfSIS and LUCAS dominate the lower range; the tail extends
      beyond 60\,g\,kg\textsuperscript{-1}.}
    \label{fig:soc-by-source}
  \end{subfigure}
  \hfill
  \begin{subfigure}[t]{0.48\textwidth}
    \centering
    \includegraphics[width=\textwidth]{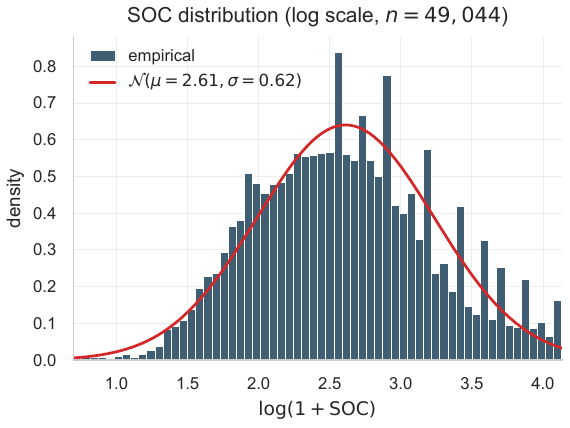}
    \caption{Log-transformed \soc{} with fitted normal
      ($\mu{=}2.62$, $\sigma{=}0.62$), confirming approximate
      log-normality.}
    \label{fig:soc-log}
  \end{subfigure}
  \caption{Distribution of \soc{} measurements across the
    49{,}044 samples.}
  \label{fig:soc-distribution}
\end{figure}

 A thorough understanding of the target variable's distribution is
fundamental to robust model training. Figure~\ref{fig:soc-distribution}
illustrates the \soc{} distribution across the source databases, alongside its
\textit{log-transformed} equivalent. Because \soc{} naturally accumulates in
specific biomes (such as \textit{peat lands}), the raw distribution is heavily
\textbf{right-skewed} (skewness = 1.61), featuring a long tail of soils with a
high carbon content, extending up to 61\,g\,kg\textsuperscript{-1}. However,
transforming the data into log space returns an approximately normal
distribution ($\mu = 2.62$, $\sigma = 0.62$). This normalization directly
motivates our use of a log-based target scaler during training, as well as the
\textbf{Huber loss} (Section~\ref{ssec:learning-objective}) to maintain
robustness against residual outliers.

\begin{table}[t]
  \centering
  \caption{Dataset summary statistics.  \soc{} values are reported
    in g\,kg\textsuperscript{-1}.}
  \label{tab:dataset-stats}
  \small
  \begin{tabular}{@{}lr@{}}
    \toprule
    Property & Value \\
    \midrule
    Total samples              & 49{,}044 \\
    Source databases            & 9 \\
    Geographic extent          & 6 continents \\
    Temporal range             & 2013--2022 \\
    \midrule
    \soc{} mean $\pm$ std      & 15.67 $\pm$ 11.38 \\
    \soc{} median              & 12.22 \\
    \soc{} range               & 1.00--61.40 \\
    \soc{} skewness            & 1.61 \\
    \midrule
    Tabular features           & 63 \\
    Sentinel-2 bands           & 11 \\
    Sentinel-1 polarisations   & 2 (VV, VH) \\
    DEM channels               & 1 \\
    \bottomrule
  \end{tabular}
\end{table}

\subsection{Data Splitting Strategy}
\label{ssec:data-splitting-strategy}

\par We split the dataset into \textbf{training} (90\%), \textbf{validation}
(5\%) and \textbf{test} (5\%) subsets using \textit{stratified random sampling}.
In order to preserve the characteristics of the target variable distribution,
\soc{} values are partitioned into \textit{quantile-based bins} and samples are
then allocated such that each split maintains a consistent representation of
these bins. This process results in approximately \textit{44{,}140} training
samples, \textit{2{,}452} validation samples and \textit{2{,}452} test samples.

\par Under a random split, nearby samples can fall in different partitions, so
the held-out metrics are best read as an upper bound on attainable accuracy. The
transductive setup (Section~\ref{ssec:problem-formulation}) keeps the same
neighbourhood structure available at training and inference time.

\par
\begin{figure}[htbp]
    \centering
    \includegraphics[width=1\textwidth]{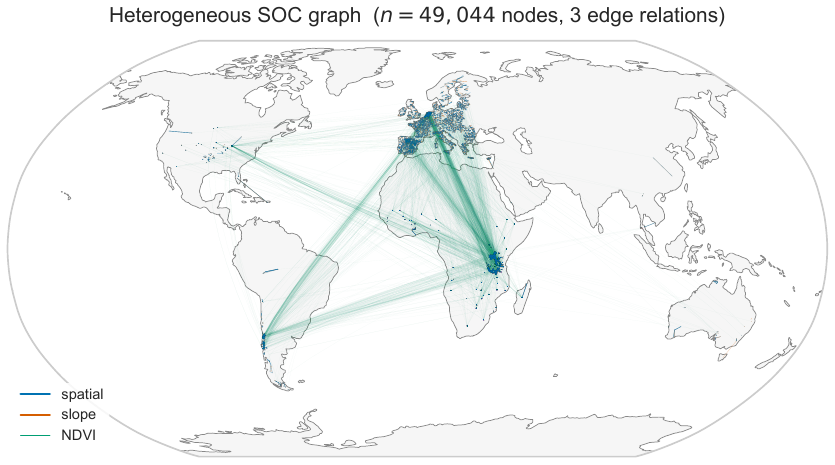}
    \caption{Global heterogeneous SOC graph projection: sample nodes connected
    by three relation types (spatial proximity, elevation similarity, NDVI
    similarity)}
    \label{fig:global_graph}
\end{figure}

 As detailed in Section \ref{ssec:problem-formulation}, the global graph
structure is built once over the entire sample pool (Figure
\ref{fig:global_graph}), with \textbf{train}, \textbf{validation} and
\textbf{test} masks applied to the node sets. The validation and test labels are
hidden from the model during training, so that the GNN's message passing
neighborhoods remain intact across all partitions (\textit{transductive setup}).

\section{Experiments}\label{sec:experiments}
\subsection{Setup}\label{ssec:setup}

\par
\begin{figure}[t]
  \centering
  \includegraphics[width=\textwidth]{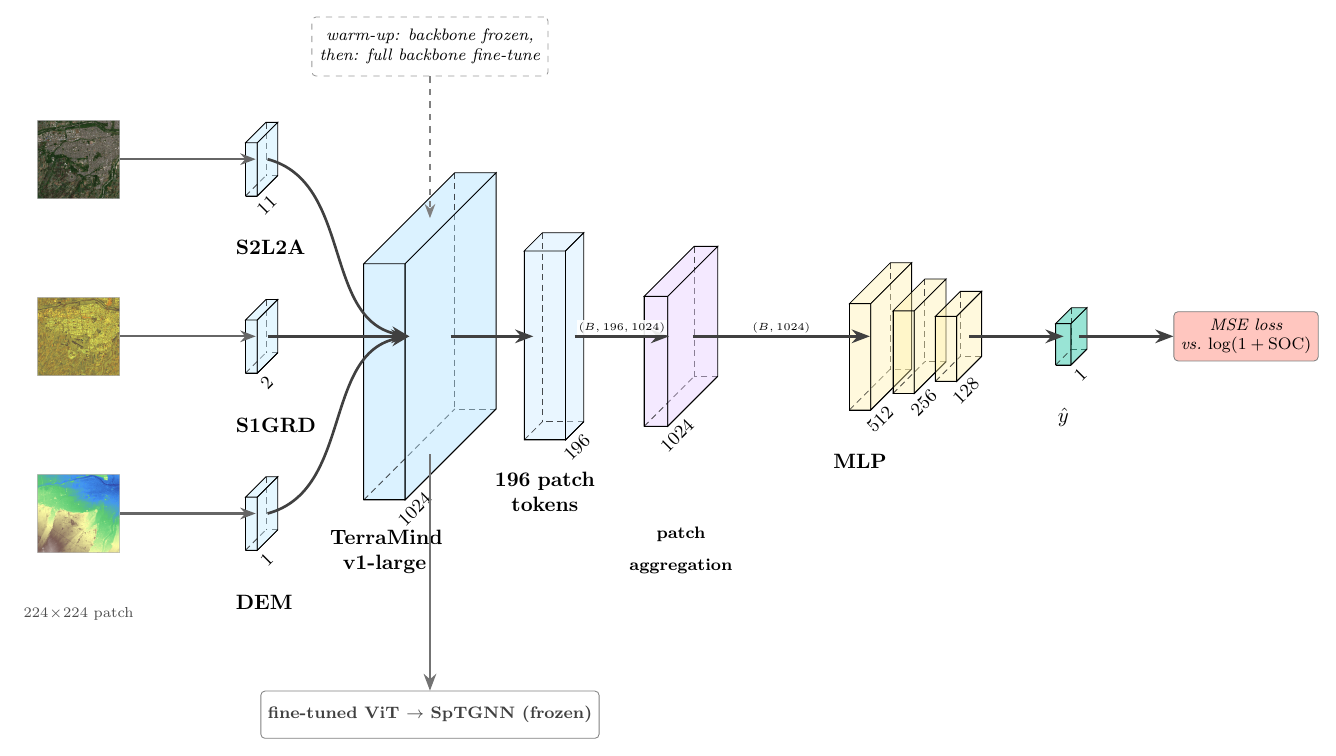}
  \caption{Per-region ViT fine-tuning pipeline. Three raster modalities
    (S2L2A, S1GRD, DEM) feed the TerraMind v1-large backbone, which is
    fine-tuned with a regression head on log-transformed SOC. The
    fine-tuned backbone is then frozen and loaded into the downstream
    \sptgnn{} model}
  \label{fig:vit-finetune}
\end{figure}

 Our experiments are carried out using three \textit{regional} sets derived
from the global dataset described in Section~\ref{sec:dataset}: \emph{Africa}
($\sim$26\,k labeled samples), \emph{Europe} ($\sim$14\,k samples) and
\emph{Global} (the whole dataset, $\sim$49\,k samples). We train our models
separately on these datasets, where they share the same architectural skeleton
(same main components described in Section \ref{sec:method}: frozen
\textit{TerraMind} ViT backbone, \textit{MoE} fusion, \textit{RGAT} stack and
prediction heads) but tuned separately. For each region we first fine-tune the
ViT model (Figure~\ref{fig:vit-finetune}, Table~\ref{tab:vit-finetune}) and then
we perform a \sptgnn{} hyperparameter search on top of the resulting frozen
backbone, resulting in the three hyperparameter configurations showcased in
Table~\ref{tab:hyperparams}.

\begin{figure}[t]
  \centering
  \includegraphics[width=\textwidth]{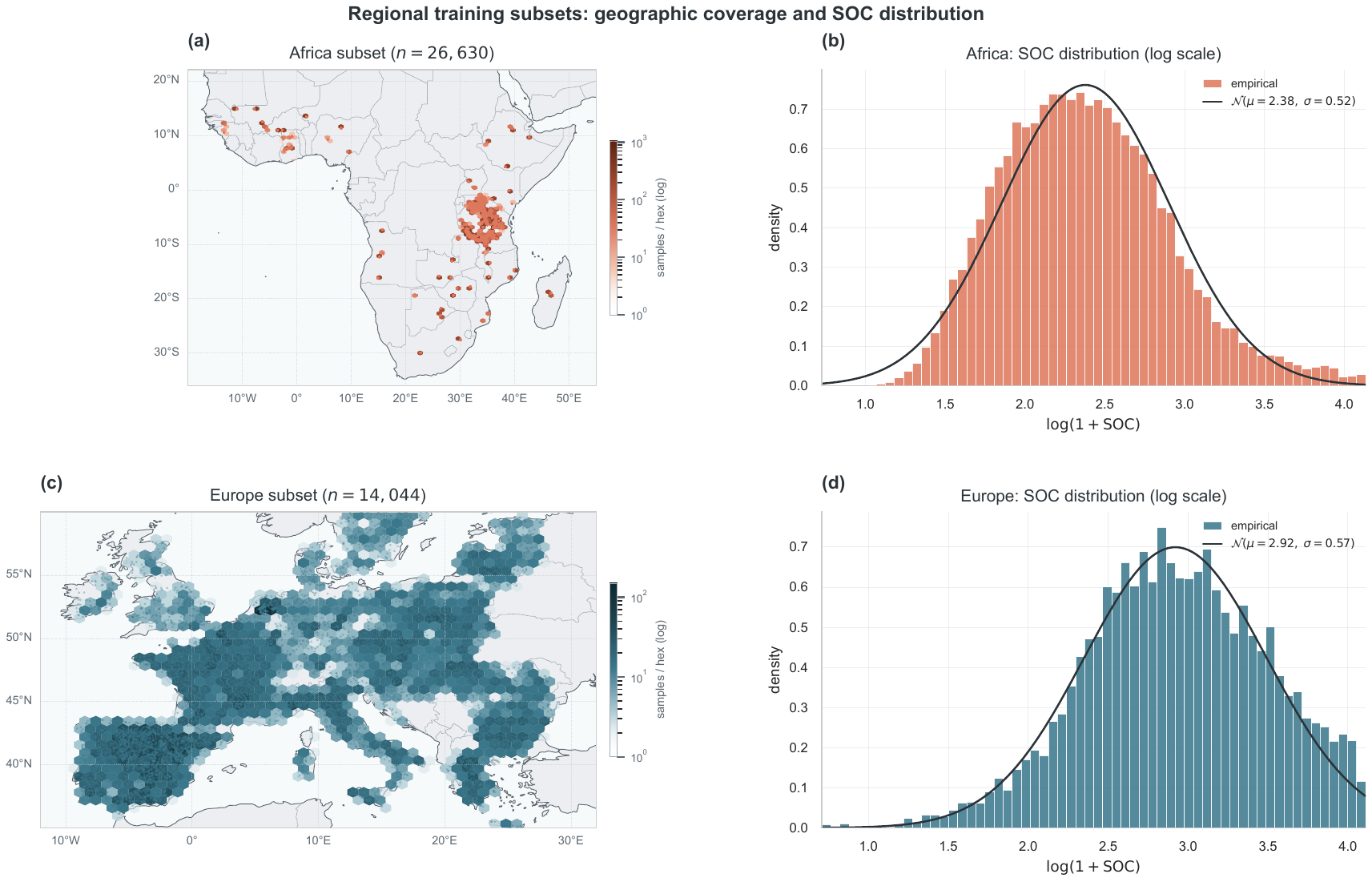}
  \caption{Regional training subsets (Africa and Europe), sample density maps on
  the left side and log-SOC distributions on the right side.}
  \label{fig:two-regions}
\end{figure}

\paragraph{Data splits}
Following the previously described splitting strategy from Section
\ref{ssec:data-splitting-strategy}, for each of the three regions, we split the
data into a 90/5/5 ratio, for train, validation and test sets. We stratify
across ten quantile bins of $\log(1+\soc{})$ using a fixed random seed. The data
graph is being built only once for each region and evaluation labels masked for
the model during the training process (Section~\ref{ssec:problem-formulation}).
The geographic distribution and density \soc{} samples for the two regions,
along with their logarithmic scale target distribution is represented in Figure
\ref{fig:two-regions}.

\paragraph{Metrics}
We evaluate our models on multiple key metrics: mean absolute error
(\textbf{MAE}), root-mean-square error (\textbf{RMSE}), mean absolute percentage
error (\textbf{MAPE}) and coefficient of determination (\textbf{$R^{2}$}). All
metrics are reported in the original \soc{} target scale, through an inverse
transformation process of the log scaler. In our case, \textbf{MAPE} is the
primary metric, as it does not depend on scale and it's therefore easier to
compare to other studies in the literature. MAE represents the average magnitude
of prediction errors and it is robust to outliers ($\geq 0$\,g\,kg$^{-1}$, lower
is better). \textbf{RMSE} is the square root of the mean squared error, which
penalizes large errors more than MAE ($\geq 0$\,g\,kg$^{-1}$, lower is better).
$R^{2}$ quantifies the proportion of variance in the dependent variable that is
predictable from the independent variable ($\leq 1$, higher is better).

\par We explain in detail the metrics used for uncertainty quantification (UQ)
in Section \ref{ssec:uq-results}, which includes: expected calibration error,
Gaussian negative log-likelihood, continuous ranked probability score, coverage
+ mean width of the 95\% \textit{Vysochanskij-Petunin} prediction interval and
the epistemic fraction ($\sigma_{\text{epistemic}}^{2} /
\sigma_{\text{total}}^{2}$).

\paragraph{Training process}
We use gradient norm clipping and a \texttt{ReduceLROnPlateau} learning rate
scheduler. The breakdown of the per-region hyperparameter search process results
can be found in Table \ref{tab:hyperparams}. Hyperparameter search process was
split on $4$ NVIDIA A100 GPUs in parallel through the \textbf{Ray Tune}
framework, while individual trials were carried out on a single NVIDIA A100 node
each.

\subsection{Cross-Region Baseline}\label{ssec:cross-region}

\par We benchmark the model performance for the three model regions. Table
\ref{tab:regions} holds the results of the previously described training and
hyperparameter tuning processes, with metrics that reflect the model's
performance on the held-out validation split for each region.

\begin{table}[t]
\centering
\small
\caption{%
  Validation metrics for the three regional \sptgnn{} instances. Each
  region was tuned independently (its own ViT fine-tune followed by its
  own \sptgnn{} hyperparameter tuning run). The three configurations share the
  same
  architectural skeleton but differ in hyperparameters
  (Table~\ref{tab:hyperparams}). The Africa instance is selected as
  the focus of the remaining experiments}
\label{tab:regions}
\begin{tabular}{lccccc}
\toprule
Region    & $n_{\text{train}}$
          & MAE $\downarrow$ & RMSE $\downarrow$
          & MAPE\,\% $\downarrow$ & $R^{2}$ $\uparrow$ \\
\midrule
\textbf{Africa} (selected)  & $\sim 25{,}200$
                            & $\mathbf{1.52}$ & $\mathbf{2.57}$
                            & $\mathbf{13.7}$ & $\mathbf{0.864}$ \\
Europe                      & $\sim 12{,}600$
                            & $6.58$ & $9.32$
                            & $37.1$          & $0.445$ \\
Global                      & $\sim 44{,}100$
                            & $4.52$ & $7.36$
                            & $29.3$          & $0.597$ \\
\bottomrule
\end{tabular}
\\[2pt]
\footnotesize
Metrics are reported on the held-out validation subset, on the
inverse-transformed (original) \soc{} scale (g\,kg$^{-1}$), at each region's
best-validation-MAPE checkpoint. These are single-model trainings using the best
configuration from the per-region hyperparameter search and are distinct from
the $5$-member deep-ensemble post-calibration test row in
Table~\ref{tab:headline}.
\end{table}

We observe the following from the region based model benchmark:

\par The \textbf{Africa} regional model achieves the highest $R^{2}$ ($0.864$
vs.\ $0.445$ for Europe and $0.597$ for Global), reflecting the broad African
\soc{} content range and tight spatial autocorrelation per-cluster (Figure
\ref{fig:density-zoom}). Europe, on the other hand, plateaus at val
$R^{2}=0.445$ and val MAPE\,$=37.1\,\%$, the LUCAS \soc{} distribution is
\emph{wider} than Africa's, so the limiting factor is not range but
heterogeneity. Lastly, the Global model is trained on the union of the Europe
and Africa sets, along with samples all across the globe, which manages to reach
val-$R^{2}=0.597$. Since data distributions vary between sources, we train
separate models per region.

\paragraph{Per-region ViT fine-tuning}
Before the \sptgnn{} training process, we fine-tune the \textit{TerraMind
v1-large backbone} on each region separately, framed as a \textbf{regression}
task. We attach a \textit{multi-layer perceptron (MLP)} regression head on top
of the tensor exported by the backbone. We feed the same three modalities as the
downstream \sptgnn{} (Sentinel-2 L2A with $11$ surface-reflectance bands,
Sentinel-1 GRD with $VV$ and $VH$ polarizations and the GLO-30 DEM) into the
TerraMind's backbone, which outputs a size \textit{$1024$ token sequence}, which
is mapped to the corresponding \textbf{log-transformed \soc{}} target through an
MLP regression head (Figure \ref{fig:vit-finetune}). We take advantage of a
\textit{warm up} period of $N_w$ epochs, keeping the backbone frozen so that the
head can adapt before unfreezing. The ViT fine-tune uses the \textbf{same
per-region 90/5/5 split} as the downstream \sptgnn{}
(Section~\ref{ssec:data-splitting-strategy}): the validation and test samples
are held out of fine-tuning, checkpoints are selected on the validation split,
and no test sample is ever seen during either the fine-tune or the \sptgnn{}
training stage.
\par The configuration of the ViT model trained on the \textbf{Africa} subset
uses \textit{Mean Squared Error (MSE)} loss, while the ones trained on
\textbf{Europe} and \textbf{Global} use \textit{Huber} loss. Head depth, warm up
length, optimizer configuration and regularization were tuned independently per
region. When we load the resulting (frozen) region based fine-tuned ViT backbone
into \sptgnn{}, the regression head is discarded. Table \ref{tab:vit-finetune}
contains the results of the region based ViT fine-tunning process, each entry
corresponds to the best performing model for that certain region, evaluated on
the validation set, taking into account $MAPE$ and $R^{2}$ metrics.

\begin{table}[t]
\centering
\small
\caption{%
  Per-region TerraMind ViT fine-tuning results. Each region's best
  validation checkpoint (selected by val-loss on the log-transformed
  target) is loaded into the corresponding SpTGNN model with frozen
  weights. We report MAPE and $R^{2}$ on the inverse-transformed
  (original) SOC scale}
\label{tab:vit-finetune}
\begin{tabular}{lcccc}
\toprule
Region & Checkpoint & val-loss $\downarrow$
       & val-MAPE\,\% $\downarrow$ & val-$R^{2}$ $\uparrow$ \\
\midrule
\textbf{Africa} & ep.\,$10$ & $0.365$ & $\mathbf{26.3}$ & $\mathbf{0.676}$ \\
Europe          & ep.\,$50$ & $\mathbf{0.163}$ & $40.8$ & $0.405$ \\
Global          & ep.\,$23$ & $0.173$ & $36.2$ & $0.512$ \\
\bottomrule
\end{tabular}
\\[2pt]
\footnotesize
val-loss is MSE on the log-transformed target (lower is better);
val-MAPE and val-$R^{2}$ are computed on the original SOC scale by
inverse-transforming the model output through the per-region target
scaler
\end{table}

\paragraph{Per-region hyperparameter search}
As previously stated, the three regional models share the same architectural
skeleton, but each was tuned separately on its own validation split, resulting
in different hyperparameter configurations. In Table \ref{tab:hyperparams} we
can see the most important parameters from a \textbf{two-step hyperparameter
tuning process}. The \textbf{first step} tunes the model with wide parameter
ranges over the most crucial configuration choices, such as GNN depth and hidden
dimensions (e.g., RGAT layer widths in
$\{[256,128],\,[256,384,192],\,[512,256,128,64]\}$), learning rate
($10^{-5}\!-\!10^{-3}$), expert type (\emph{cross-gated},  \emph{concat},
\emph{additive}), MoE fusion dimension ($\{256, 512, 1024\}$), number of experts
($\{4, 8, 16\}$), attention mechanism, dropout, weight decay and batch size.
Each configuration is trained on the regional training splits and ranked by
validation MAPE and $R^{2}$.

\par For the \textbf{second step} we analyze the top quartile of trials from
step one and we identify the resulting values that consistently appear in the
best performing configurations. We then tighten the search space around the
identified hotspots, using more granular choices and pruning categorical choices
(e.g., expert type fixed to \emph{cross-gated} and learning rate restricted to
$\bigl[1{\times}10^{-4},\,3{\times}10^{-4}\bigr]$).

\par The Table \ref{tab:hyperparams} reveals that the Africa \sptgnn{} model
converges on the deepest GNN configuration, with a \emph{within-relation
multiplicative} attention mode, while the Europe model achieves best performance
using a shallower three layer GNN with \emph{additive} attention. The Global
configuration uses a two layer GNN with \emph{across-relation} additive
attention. Components that are more \textit{common} are summarized in the second
half the table in order to keep the comparison easy to read.

\begin{table}[t]
\centering
\small
\caption{%
  Most important hyperparameters for each regional SpTGNN instance,
  selected from the per-region best configurations. The upper block
  lists the parameters that differ across regions; the lower block
  lists settings that are shared by all three.}
\label{tab:hyperparams}
\begin{tabular}{lccc}
\toprule
                                 & \textbf{Africa}                & \textbf{Europe}                  & \textbf{Global}                  \\
\midrule
\multicolumn{4}{l}{\emph{Optimisation}} \\
Learning rate (Adam)             & $1.95\!\times\!10^{-4}$        & $2.19\!\times\!10^{-4}$          & $5.27\!\times\!10^{-4}$          \\
Weight decay                     & $2.79\!\times\!10^{-5}$        & $1.53\!\times\!10^{-4}$          & $5.29\!\times\!10^{-5}$          \\
Gradient-clip $\lVert g\rVert_2$ & $0.5$                          & $0.5$                            & $1.87$                           \\
Epochs (max)                     & $100$                          & $100$                            & $50$                             \\
Early-stop patience              & $30$                           & $800$                            & $800$                            \\
LR-scheduler patience            & $10$                           & $100$                            & $100$                            \\
Batch size                       & $32$                           & $16$                             & $16$                             \\
Auxiliary-loss weight $\lambda_\text{aux}$ & $5.3\!\times\!10^{-5}$  & $8.8\!\times\!10^{-5}$           & $4.9\!\times\!10^{-5}$           \\
\midrule
\multicolumn{4}{l}{\emph{GNN architecture}} \\
RGAT hidden dims                 & $[512, 256, 128, 64]$          & $[256, 384, 192]$                & $[256, 128]$                     \\
Number of bases                  & $4$                            & $3$                              & --- \\
Attention mechanism              & within-relation                & within-relation                  & across-relation                  \\
Attention mode                   & multiplicative                 & additive                         & additive                         \\
Negative slope (LeakyReLU)       & $0.106$                        & $0.100$                          & $0.200$                          \\
Dropout                          & $0.286$                        & $0.341$                          & $0.236$                          \\
\midrule
\multicolumn{4}{l}{\emph{Mixture-of-Experts fusion}} \\
Number of experts                & $8$                            & $16$                             & $8$                              \\
Expert type                      & cross-gated                    & cross-gated                      & concat                           \\
Fusion dim                       & $512$                          & $1024$                           & $1024$                           \\
Top-$k$ routing                  & dense                          & $4$                              & $4$                              \\
Load-balance weight              & $1.95\!\times\!10^{-2}$        & $3.38\!\times\!10^{-3}$          & $6.19\!\times\!10^{-2}$          \\
\midrule
\multicolumn{4}{l}{\emph{Pre / Post MLPs}} \\
Pre-MLP hidden $\to$ output      & $[256, 128] \to 256$           & $[1024, 512] \to 512$            & $[1024] \to 1024$                \\
Pre-MLP dropout                  & $0.161$                        & $0.176$                          & $0.225$                          \\
Post-MLP hidden $\to$ output     & $[256, 128] \to 128$           & $[512, 256] \to 128$             & $[256, 128] \to 32$              \\
Post-MLP dropout                 & $0.10$                         & $0.27$                           & $0.27$                           \\
\midrule
\multicolumn{4}{l}{\emph{Graph construction}} \\
HDBSCAN min cluster size         & $8$                            & $2$                              & $2$                              \\
Adaptive-$k$ neighbours          & $10$                           & $5$                              & $5$                              \\
\midrule
\multicolumn{4}{l}{\emph{Encoders (ViT, time, positional) and graph relations}} \\
\multicolumn{4}{l}{\quad ViT backbone: TerraMind v1-large (frozen), CLS-token patch-merge transformer.} \\
\multicolumn{4}{l}{\quad CLS-merge (Tx, layers / heads / ff): Africa $2/4/1024$, Europe $3/8/1024$, Global $2/4/512$.} \\
\multicolumn{4}{l}{\quad Modality merge: Africa, Europe = mean; Global = max.} \\
\multicolumn{4}{l}{\quad Time encoder: $4$ Fourier frequencies, output dim $8$ (all regions).} \\
\multicolumn{4}{l}{\quad Positional encoder: $10$ scales, $\sigma\in[0.1,10]$, output dim $16$ (all regions).} \\
\multicolumn{4}{l}{\quad Spatial / NDVI / elevation graph relations with $k_{\text{spatial}}{=}5$, $k_{\text{NDVI}}{=}3$,} \\
\multicolumn{4}{l}{\quad max distance $2{,}000$\,km, slope scale $30$ (all regions).} \\
\bottomrule
\end{tabular}
\end{table}

\paragraph{Benchmark subset}
As our \textbf{Africa} region model has the strongest results in our evaluation
benchmarks, for the rest of our study (ablations, loss-function comparisons and
uncertainty calibration in Sections~\ref{ssec:predictive}-\ref{ssec:uq-results})
we evaluate strictly on this subset. This presents a strong signal to noise
ratio, so that we can clearly quantify the removal of certain model components.
The \textbf{AfSIS} subset, which mostly comprises the Africa training set, being
a homogeneous set eliminates the need to account for distribution shifts,
meaning that the performance gaps can be strictly attributable to the model
design. Being a smaller subset ($n_{\text{train}}=25{,}180$,
$n_{\text{val}}=768$, $n_{\text{test}}=682$), our experiments take considerably
less time to finish, we approximate a mean of 4.5 hours for a training run on an
A100 GPU.

\subsection{Uncertainty Quantification}\label{ssec:uq-results}

\par We set up \textbf{two $5$-member ensembles} in order to compute the
calibration and uncertainty decomposition of our model, specifically
(\textbf{MSE $+$ $\lambda$NLL}, $\lambda=0.05$, $30$-epoch \textit{warm start)}
and \textbf{$\beta$-NLL} ($\beta=0.5$, NLL-as-primary, \textbf{no} \textit{warm
start}), as previously described in Section~\ref{ssec:uncertainty}.

\paragraph{Calibration before and after temperature scaling}

\begin{figure}[t]
  \centering
  \includegraphics[width=0.6\textwidth]{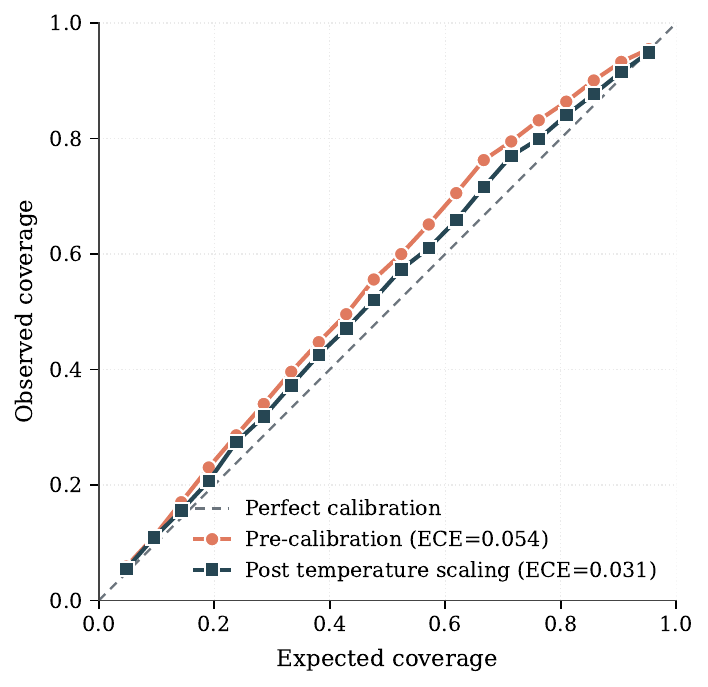}
  \caption{Reliability diagrams (predicted quantile vs.\ empirical
    coverage) for the hybrid and $\beta$-NLL ensembles on the
    Africa test split, before and after post-hoc temperature scaling.}
  \label{fig:reliability}
\end{figure}

Both ensemble configurations show uncertainty miscalibration.
Table~\ref{tab:calibration} reports the \textit{Africa} test-split calibration
results before and after applying \textbf{temperature scaling} with a
\textbf{scalar parameter} fitted on the validation set. Before calibration, both
\sptgnn{} ensembles over-cover the nominal interval ($\mathrm{ECE}=0.054$,
Cov@95\,$=0.981$) and the $\beta$-NLL ensemble over-covers even more strongly
($\mathrm{ECE}=0.085$, Cov@95\,$=0.994$). Temperature scaling improves
calibration for both models, which reduces ECE by $1.7$-$3.3\times$ and brings
the two ensembles to comparable quality ($\mathrm{ECE}_{\text{post}} \approx
0.026$-$0.031$), without affecting the point predictions.
Figure~\ref{fig:reliability} shows the corresponding reliability diagram for the
\sptgnn{} ensemble.

\begin{table}[t]
\centering
\small
\caption{
  Calibration metrics on the Africa test split before and after post-hoc
  temperature scaling. Brackets are paired bootstrap $95\,\%$ CIs over
  $1{,}000$ test resamples. Cov@95 / MPIW@95 use distribution free
  \textit{Vysochanskij-Petunin} intervals; ECE / NLL / CRPS evaluate the model's
  Gaussian predictive}
\label{tab:calibration}
\begin{tabular}{l@{\hskip 8pt}cc@{\hskip 8pt}cc}
\toprule
 & \multicolumn{2}{c}{Production (hybrid)}
 & \multicolumn{2}{c}{$\beta$-NLL ($\beta{=}0.5$)} \\
\cmidrule(lr){2-3}\cmidrule(l){4-5}
Metric          & Pre-cal               & Post-cal ($T{=}0.92$)
                & Pre-cal               & Post-cal ($T{=}0.83$) \\
\midrule
ECE $\downarrow$           & $0.054$\,{\scriptsize[0.034, 0.077]} & $0.031$\,{\scriptsize[0.013, 0.054]}
                           & $0.085$\,{\scriptsize[0.063, 0.105]} & $0.026$\,{\scriptsize[0.011, 0.048]} \\
NLL  $\downarrow$          & $-0.194$ & $-0.183$ & $-0.189$ & $-0.206$ \\
CRPS $\downarrow$          & $0.259$  & $0.257$  & $0.273$  & $0.269$  \\
Cov@95 ($\rightarrow 0.95$)& $0.981$  & $0.981$  & $0.994$  & $0.990$  \\
MPIW@95 $\downarrow$       & $3.23$   & $2.98$   & $3.48$   & $2.89$   \\
\bottomrule
\end{tabular}
\end{table}

\paragraph{Uncertainty decomposition}

\begin{figure}
  \centering
  \includegraphics[width=0.55\textwidth]{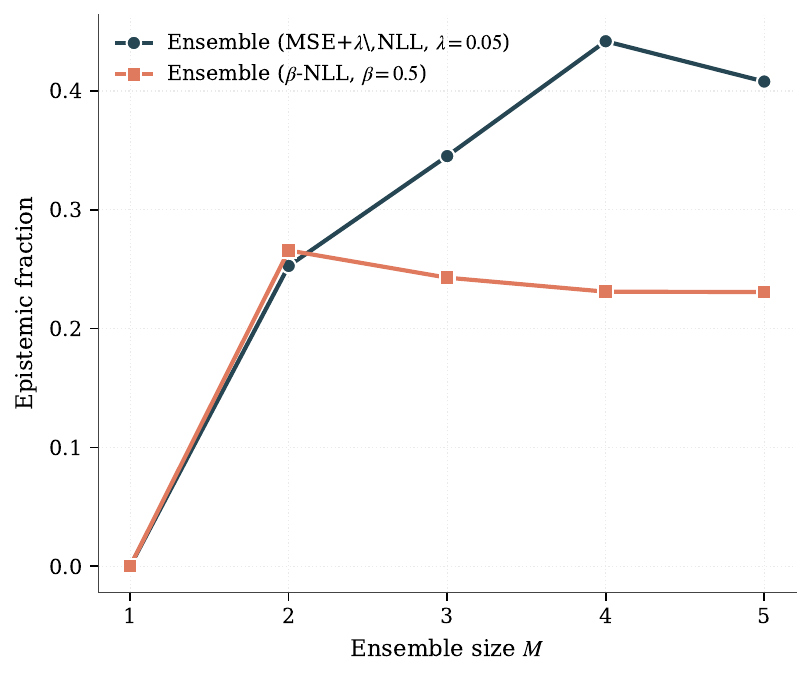}
  \caption{Epistemic fraction
    $\sigma_{\text{epistemic}}^{2}/\sigma_{\text{total}}^{2}$ as a
    function of ensemble size $M$. Both ensembles are zero at $M{=}1$
    by construction (single member, no between-member variance). The
    SpTGNN hybrid keeps growing through $M{=}4$ as ensembling rescues
    the $\sigma$-collapse regime, while $\beta$-NLL plateaus near
    $0.23$.}
  \label{fig:epistemic-fraction}
\end{figure}

In Table~\ref{tab:uq-decomposition} we split the post calibration prediction
variance in \textbf{aleatoric} and \textbf{epistemic} uncertainty. The model
ensemble determines that $48\,\%$ of the total variance is \textbf{epistemic},
compared to $34\,\%$ for the $\beta$-NLL ensemble, with the gap widening
(Figure~\ref{fig:epistemic-fraction}). These results show us that almost half of
the predicted uncertainty is attributed to \emph{ensemble member disagreement}.
Whereas the hybrid $\lambda$NLL ensemble setup encourages the members to produce
\textbf{conservative} aleatoric uncertainty estimates (warm up and small
$\lambda=0.05$).

\begin{table}[t]
\centering
\small
\caption{%
  Uncertainty decomposition on the Africa test split, post-calibration.
  Brackets are bootstrap $95\,\%$ CIs.}
\label{tab:uq-decomposition}
\begin{tabular}{lcc}
\toprule
                                  & SpTGNN                  & $\beta$-NLL \\
\midrule
Aleatoric variance (raw)          & $0.176$                 & $0.142$ \\
Epistemic variance (raw)          & $0.126$\,{\scriptsize[0.112, 0.143]} & $0.108$\,{\scriptsize[0.095, 0.122]} \\
Total variance (raw)              & $0.299$                 & $0.249$ \\
Epistemic fraction                & $\mathbf{0.480}$\,{\scriptsize[0.452, 0.505]} & $0.335$\,{\scriptsize[0.317, 0.353]} \\
Pearson $r(|\text{err}|, \sigma)$ & $0.292$                 & $0.314$ \\
Spearman $r(|\text{err}|, \sigma)$& $0.262$                 & $0.268$ \\
\bottomrule
\end{tabular}
\end{table}

\paragraph{Sensitivity to ensemble size}

\begin{figure}[t]
  \centering
  \includegraphics[width=0.49\textwidth]{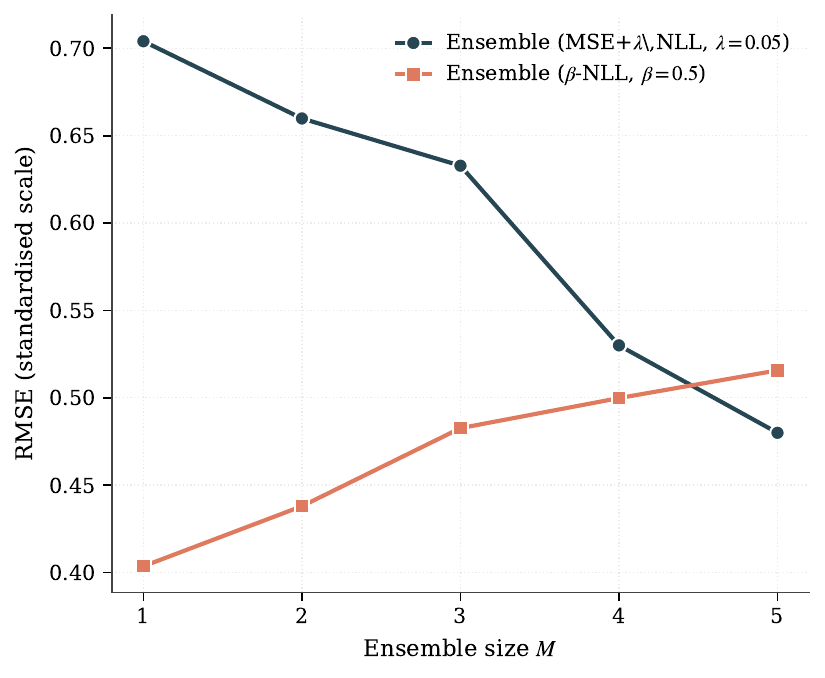}\hfill
  \includegraphics[width=0.49\textwidth]{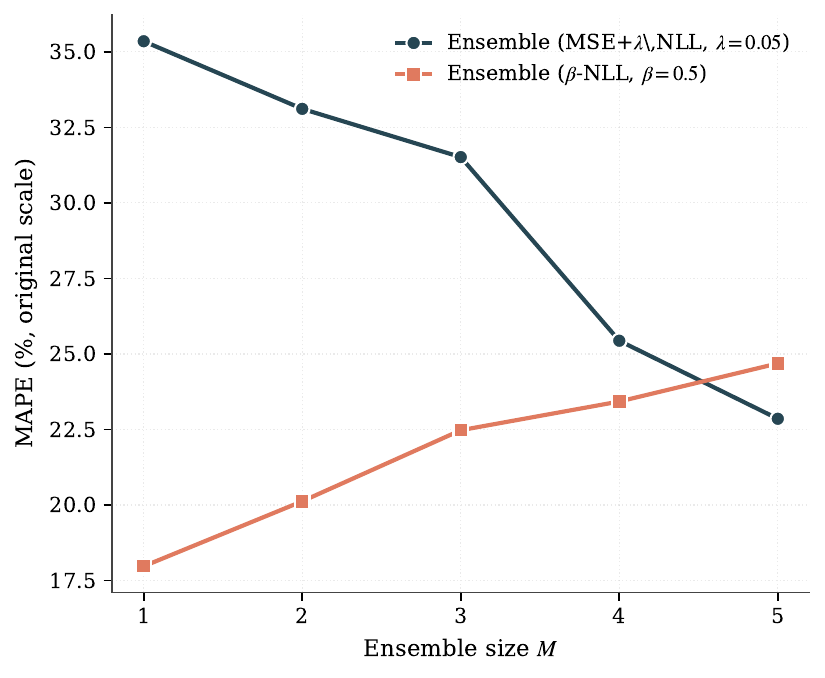}\\[2pt]
  \includegraphics[width=0.49\textwidth]{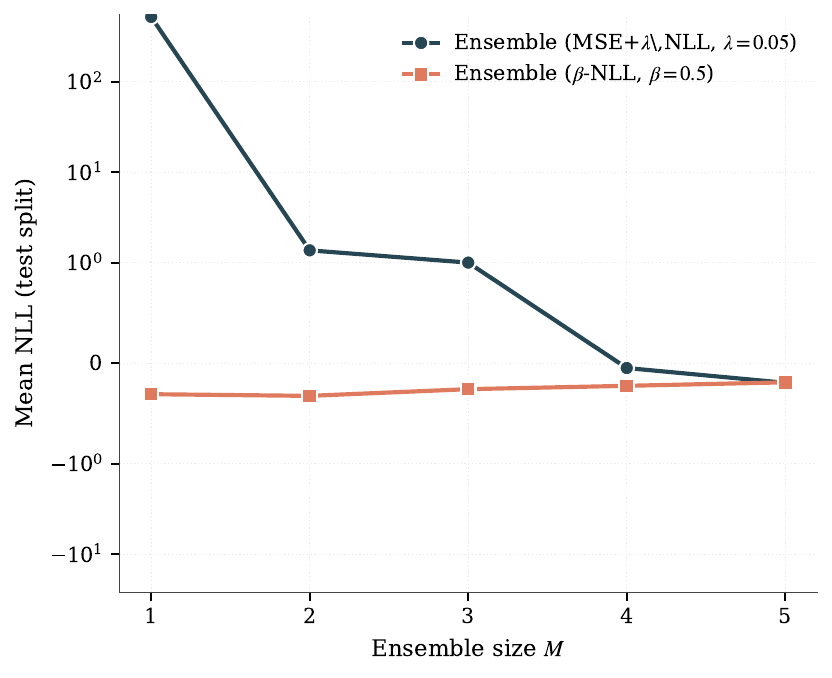}\hfill
  \includegraphics[width=0.49\textwidth]{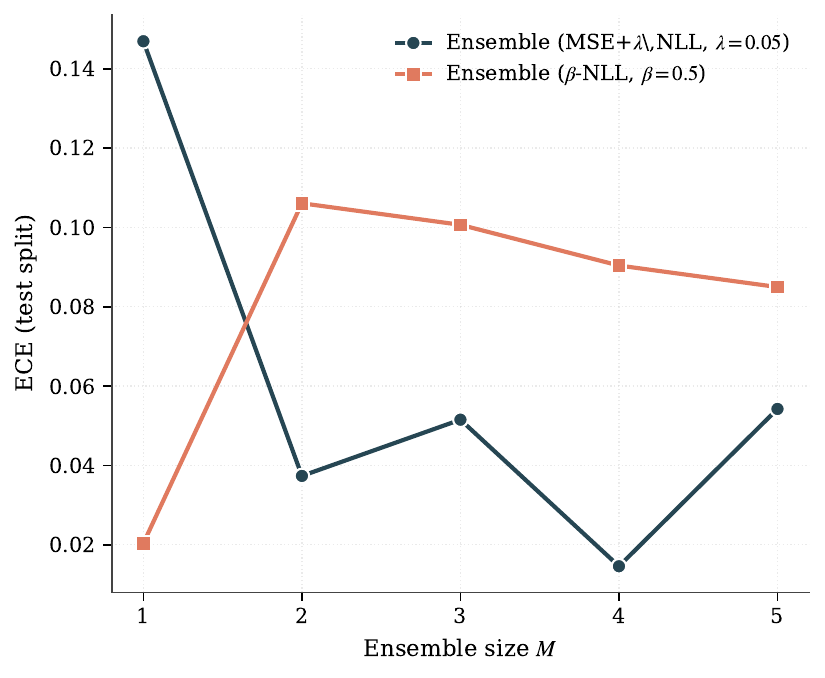}
  \caption{Effect of ensemble size $M$ on accuracy
    (top: RMSE on the standardised scale, MAPE on the original scale)
    and on calibration (bottom: NLL on a symlog scale, ECE), for the
    SpTGNN hybrid (dark) and $\beta$-NLL (orange) losses on the
    Africa test split. The two losses produce opposite trends:
    the SpTGNN hybrid improves monotonically with $M$, escaping
    the heteroscedastic single-model $\sigma$-collapse, while
    $\beta$-NLL is already calibrated at $M{=}1$ and degrades as $M$
    grows.}
  \label{fig:ensemble-size}
\end{figure}

We measure the impact of the ensemble size, by iteratively evaluating different
subsets formed from the initial $5$ members of the ensemble, without the need
for re-training (Table~\ref{tab:ensemble-size} and
Figure~\ref{fig:ensemble-size}). As the ensemble grows in members, the two
losses behave differently. When we are using an ensemble of size $1$ and the
hybrid \sptgnn{} loss, the model often predicts unrealistically small
uncertainty values ($\sigma \to 0$) on difficult samples, resulting in poor NLL
($521.9$) and low coverage (Cov@95\,$=0.55$). As we add more members, this issue
gets resolved because ensemble disagreement adds uncertainty, as per Eq.
\ref{eq:ensemble}. The gradual increase from $1$ to $5$ members improves RMSE
from $0.70$ to $0.48$, while NLL almost stabilizes near its optimum ($\sim
-0.19$).

\par $\beta$-NLL shows the opposite, an ensemble of size $1$ is already
well-calibrated ($\mathrm{NLL}=-0.31$, $\mathrm{ECE}=0.020$, Cov@95,$=0.98$) and
adding more members increases the total predicted uncertainty lowers the RMSE
and calibration.

\begin{table}[t]
\centering
\small
\caption{%
  Calibration and accuracy as a function of ensemble size $M$, evaluated
  on the Africa test split with no retraining (post-hoc member subsetting).}
\label{tab:ensemble-size}
\begin{tabular}{l@{\hskip 6pt}rrrrr}
\toprule
                       & $M{=}1$ & $M{=}2$ & $M{=}3$ & $M{=}4$ & $M{=}5$ \\
\midrule
\multicolumn{6}{l}{\emph{SpTGNN (MSE $+\lambda$NLL hybrid)}}\\
RMSE $\downarrow$              & $0.704$ & $0.660$ & $0.633$ & $0.530$ & $\mathbf{0.480}$ \\
NLL $\downarrow$               & $521.9$ & $1.37$  & $1.00$  & $-0.05$ & $\mathbf{-0.19}$ \\
ECE $\downarrow$               & $0.147$ & $0.037$ & $0.052$ & $\mathbf{0.015}$ & $0.054$ \\
Cov@95 ($\rightarrow 0.95$)    & $0.553$ & $0.900$ & $0.919$ & $\mathbf{0.978}$ & $0.981$ \\
Epistemic frac.                & $0.000$ & $0.253$ & $0.345$ & $0.442$ & $0.408$ \\
\midrule
\multicolumn{6}{l}{\emph{$\beta$-NLL}}\\
RMSE $\downarrow$              & $\mathbf{0.404}$ & $0.438$ & $0.483$ & $0.500$ & $0.516$ \\
NLL $\downarrow$               & $\mathbf{-0.31}$ & $-0.32$ & $-0.26$ & $-0.22$ & $-0.19$ \\
ECE $\downarrow$               & $\mathbf{0.020}$ & $0.106$ & $0.101$ & $0.090$ & $0.085$ \\
Cov@95 ($\rightarrow 0.95$)    & $0.984$ & $0.994$ & $0.996$ & $0.994$ & $0.994$ \\
Epistemic frac.                & $0.000$ & $0.266$ & $0.243$ & $0.231$ & $0.231$ \\
\bottomrule
\end{tabular}
\end{table}

\subsection{Predictive Performance}\label{ssec:predictive}

\par In Table \ref{tab:headline} we showcase the predictive performance on the
held out Africa test subset for the \textit{\sptgnn{} ensemble}, trained using
MSE $+$ $\lambda$NLL ($\lambda = 0.05$) with a $30$-epoch warm-start phase,
which is an improvement when compared to the XGBoost baseline. RMSE is reduced
by approximately $8.5\,\%$ ($3.51$ compared to $3.84$), lowering MAPE from
$25.2\,\%$ to $22.9\,\%$.

\par The $\beta$-NLL ensemble is worse than \sptgnn{} on both point metrics: its
RMSE is higher by $\sim$$12\,\%$ ($3.93$ vs.\ $3.51$) and its MAPE is also
higher ($24.7\,\%$ vs.\ $22.9\,\%$). The $\beta$-NLL configuration is retained
not for point accuracy but for its uncertainty behaviour, which we break down in
depth in Section~\ref{ssec:uq-results}.

\begin{table}[t]
\centering
\small
\caption{%
  Point-prediction performance on the Africa test split ($n_{\text{test}}=682$,
  units: g\,kg$^{-1}$). Bold marks the best of each column. \sptgnn{}
  ensemble error bars are $1{,}000$-resample bootstrap $95\,\%$ CIs.}
\label{tab:headline}
\begin{tabular}{lcccccc}
\toprule
Method & $n_{\text{test}}$ & MAE $\downarrow$ & RMSE $\downarrow$
       & MAPE\,\% $\downarrow$ & $R^{2}$ $\uparrow$ & UQ \\
\midrule
XGBoost (covariates only)         & 682 & $2.52$           & $3.84$           & $25.2$          & $0.715$           & --- \\
$\beta$-NLL ensemble ($M{=}5$)    & 682 & $2.42$           & $3.93$           & $24.7$          & $0.700$           & \checkmark \\
\textbf{\sptgnn{} ensemble} ($M{=}5$)
   & 682 & $\mathbf{2.27}$ & $\mathbf{3.51 \pm 0.48}$ & $\mathbf{22.9}$ & $\mathbf{0.762}$ & \checkmark \\
\bottomrule
\end{tabular}
\end{table}

\par
\begin{figure}[t]
  \centering
  \includegraphics[width=\textwidth]{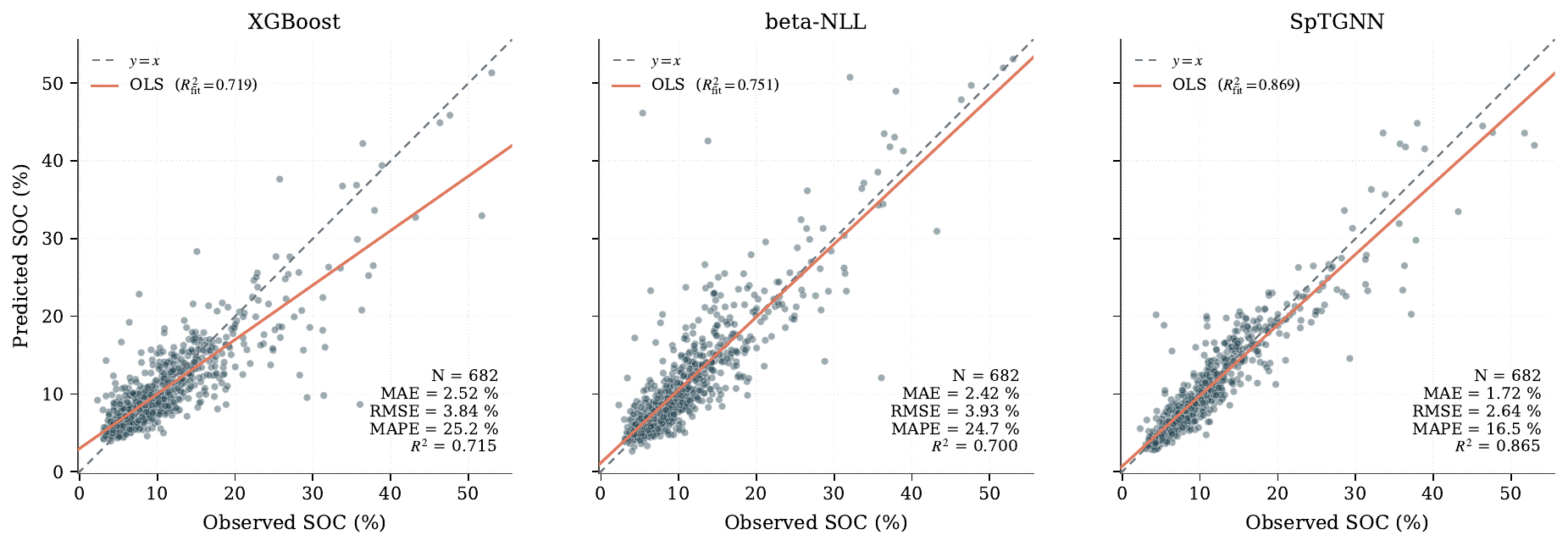}
  \caption{Predicted vs.\ true \soc{} on the Africa test split for the
    three configurations. Diagonal is the $1{:}1$ line; colour encodes point
    density}
  \label{fig:scatter-3panel}
\end{figure}

 The results of our evaluation can be seen in
Figure~\ref{fig:scatter-3panel}, with  predicted vs.\ true \soc{}, corresponding
to all three rows from Table \ref{tab:headline}. We can see that the \sptgnn{}
ensemble's residuals are more concentrated around the $1{:}1$ diagonal.

\subsection{Architecture Ablations}\label{ssec:ablations}

\par We compare our Africa \sptgnn{} model to a simple tabular model (XGBoost)
and three different \sptgnn{} architecture ablations:
\begin{itemize}
    \item \textbf{XGBoost (with only covariates)}: gradient-boosted trees
    \citep{chen_xgboost_2016} trained on the $63$ environmental covariates, with
    \textbf{no imagery} and \textbf{no graph} ($n_{\text{est}}=500$, depth $=8$,
    $\eta=0.05$)
    \item \textbf{Spatial graph}: \sptgnn{} with the heterogeneous graph
    collapsed to a single edge type (geographic $k$-NN only based on lat/lon)
    \item \textbf{Concatenation based fusion}: \sptgnn{} with the
    Mixture-of-Experts (MoE) replaced by feature concatenation followed by an
    MLP
    \item \textbf{ViT disabled}: \sptgnn{} with the \textbf{TerraMind ViT}
    embedding stream not joining at the fusion layer, leaving only tabular and
    positional inputs
\end{itemize}

\par All of these models are trained on the Africa sub set and evaluated on the
same \textbf{validation split} and metrics are reported in the inverse
transformed (original) \soc{} scale. The \textit{Africa} trained \sptgnn{} is
included in Table \ref{tab:ablations} as a reference as our best configuration.

\begin{table}[t]
\centering
\small
\caption{
  Architecture ablations on the Africa instance, all trained on the
  same data split and evaluated on the validation split, on the
  inverse-transformed (original) SOC scale (g\,kg$^{-1}$). Each row is
  a single-model checkpoint trained under the shared ablation protocol
  (same seed, schedule and budget for all four variants), so the
  \textbf{SpTGNN (full)} row is a controlled re-run of the Africa
  configuration and is slightly below the cherry-picked best-HPO
  checkpoint in Table~\ref{tab:regions} ($R^{2}{=}0.864$); the gap
  reflects HPO selection variance, not a different model.}
\label{tab:ablations}
\begin{tabular}{lcccc}
\toprule
Variant                        & MAE $\downarrow$ & RMSE $\downarrow$ & MAPE\,\% $\downarrow$ & $R^{2}$ $\uparrow$ \\
\midrule
\textbf{SpTGNN (full)}         & $\mathbf{1.72}$  & $\mathbf{2.64}$   & $\mathbf{16.5}$       & $\mathbf{0.865}$    \\[2pt]
Spatial-only graph             & $3.04$           & $5.02$            & $28.2$                & $0.479$             \\
Concat fusion                  & $2.80$           & $4.55$            & $26.1$                & $0.571$             \\
ViT disabled                   & $2.84$           & $4.31$            & $28.4$                & $0.611$             \\
\bottomrule
\end{tabular}
\end{table}

\par Dropping the \textbf{NDVI} and \textbf{elevation} relational edges reduces
performance, with an $R^{2}$ loss of nearly $0.4$ on validation (RMSE $5.02$,
$R^{2}=0.479$, MAPE $28.2\,\%$), supporting the use of heterogeneous instead of
the positionally encoded homogeneous graph. The replacement of
Mixture-of-Experts (MoE) fusion with a simple concatenation decreases $R^{2}$ by
$\sim$$0.29$ (RMSE $4.55$, MAPE $26.1\,\%$), the gap being smaller than the
graph ablation, but reflecting the lack of modality-specific interactions.
Lastly, removing TerraMind ViT as input sees a difference in $R^{2}$ by
$\sim$$0.25$ (RMSE $4.31$, MAPE $28.4\,\%$), losing the local area
representation offered by the foundation model. Altogether, the ablation study
results show that each major architectural component contributes meaningfully to
the overall performance of the \sptgnn{} model on the \textit{Africa} dataset.

\subsection{Diagnostics}\label{ssec:diagnostics}

\paragraph{Gaussian-Assumption Checks}

The \textbf{Gaussian negative log-likelihood} objective assumes the targets are
conditionally independent and Gaussian, $y_i \mid x_i \sim
\mathcal{N}(\hat{y}_i,\sigma_i^2)$, with the variance set by the predicted
heteroscedastic uncertainty $\sigma_i^2$. We evaluate this, with the report
shown in Table~\ref{tab:diagnostics} for the three tests on the post-calibration
residuals for both ensembles. Both ensembles \emph{fail} the
\textit{Shapiro-Wilk} \citep{eb32428d-e089-3d0c-8541-5f3e8f273532} normality
test, which indicates a statistically significant deviations from Gaussianity.
The \textit{Breusch-Pagan} \citep{59931fb7-1472-3118-95f0-a6dba150b325}
homoscedasticity test, suggests that variance remains correlated with either the
predicted uncertainty or the predicted mean, meaning that the heteroscedastic
structure is not fully captured by the learned variance head.

\begin{table}[t]
\centering
\small
\caption{Goodness-of-fit diagnostics on standardised Africa test
  residuals $z_i = (y_i - \mu_i)/\sigma_i$ for the two ensembles.
  \textbf{Shapiro-Wilk} and \textbf{Breusch-Pagan} test statistics are scale
  independent
  under temperature scaling; the moments of $z$ are reported
  post-calibration. Target values under correct specification:
  $\mathrm{mean}(z)=0$, $\mathrm{var}(z)=1$.}
\label{tab:diagnostics}
\begin{tabular}{lcc}
\toprule
Statistic                                            & SpTGNN              & $\beta$-NLL ($\beta{=}0.5$) \\
\midrule
Shapiro-Wilk $W$ \,/\, $p$                          & $0.934 / \!\sim\!10^{-16}$ & $0.947 / \!\sim\!10^{-15}$ \\
Breusch--Pagan $z^{2}\!\sim\!\sigma$ \,/\, $p$        & $9.94 / 0.0016$            & $0.26 / 0.609$ \\
Breusch--Pagan $z^{2}\!\sim\!\mu$    \,/\, $p$        & $0.42 / 0.516$             & $10.14 / 0.0015$ \\[2pt]
$\mathrm{mean}(z)$                                   & $-0.139$                   & $-0.094$ \\
$\mathrm{var}(z)$                                    & $1.212$                    & $1.083$  \\
\bottomrule
\end{tabular}
\end{table}

\paragraph{Mean and variance}

Table \ref{tab:diagnostics} shows that the residuals are close to the targets,
$\mathrm{mean}(z)=-0.14$, $\mathrm{var}(z)=1.21$ and $\beta$-NLL $-0.09$ and
$1.08$ (targets $0$ and $1$). Both models slightly over predict, with \sptgnn{}
deviating more from the ground truth.

\paragraph{Heteroscedasticity of the standardized residuals}

\textbf{Breusch-Pagan} tests if the variance of the error depends on a
regressor. In our \sptgnn{}, we apply the test to $z^2$ against the standard
deviation $\sigma$ and predicted mean $\mu$. Under heteroscedastic Gaussian,
$z^2$ should be independent of both. For SpTGNN, $z^2$ depends on $\sigma$
($p=0.0016$) but not on $\mu$ ($p=0.516$), the magnitude of the standardized
residual varies across uncertainty, meaning the $\sigma$ head \textit{over} or
\textit{under}-estimates variance in some regions of its own output range, even
though the residual spread is independent of the prediction itself. For
$\beta$-NLL, the pattern is reversed, $z^2$ is independent of $\sigma$
($p=0.609$) but depends on $\mu$ ($p=0.0015$), so the $\sigma$ head is well
scaled along its range but leaves residual variance that varies with the
prediction. The hybrid objective learns a weaker variance head, whereas
$\beta$-NLL sharpens variance estimation at the cost of mean dependent
heteroscedasticity.

\paragraph{Shape of the predictive distribution}

\textbf{Shapiro-Wilk} rejects Gaussianity for both ensembles ($p\approx10^{-16}$
and $10^{-15}$). Nevertheless, the post-calibration coverage goes over the
intended $95\%$ level ($0.981$ for SpTGNN and $0.990$ for $\beta$-NLL), this
indicates that the predictive distributions are \textbf{too conservative},
rather than \textbf{over confident}.

\paragraph{Reported metrics}

Because the Gaussian assumption is mis-satisfied, we report both
Gaussian-likelihood metrics (NLL) and the calibration-and-sharpness metrics
(CRPS, ECE) computed under our Gaussian predictive, alongside distribution-free
coverage metrics (Cov@95, MPIW@95) derived from Vysochanskij–Petunin bounds
\citep{priore2023chanceconstrainedstochasticoptimal}, which require only finite
variance and unimodality. Both provide consistent conclusions for our
experiments, see Section~\ref{sec:discussion}.

\section{Discussion}\label{sec:discussion}
\par The experiments described in Section \ref{sec:experiments} support the main
claims of this paper. In this section we go through the assumptions, caveats and
open problems that limit these results.

\subsection{Limitations}
\label{ssec:limitations}

\paragraph{Gaussian-likelihood violation}

The \textit{heteroscedastic Gaussian NLL} assumes $z_i = (y_i - \mu_i)/\sigma_i
\sim \mathcal{N}(0,1)$. Through the tests performed in
Section~\ref{ssec:diagnostics} (\textit{Shapiro-Wilk} and
\textit{Breusch-Pagan}), we reject Gaussianity and reveal that the two ensembles
show different residual heteroscedasticity patterns. The interval of predictions
is slightly wider than needed and we believe that the impact on practical use is
small.

\paragraph{Per-sample independence in the uncertainty model}

The model ensemble with temperature scaling pipeline only produces the
\textit{predicted standard deviation} for each sample, independently. Whereas,
in reality, samples are correlated by proximity and location, leading to
correlations in errors. When predictions are aggregated over a specific region,
the calculated uncertainty should not be the sum of the per-sample
uncertainties, but instead, extending \sptgnn{}'s predictions with a
\textit{Gaussian process residual} or a \textit{kriged correction} fitted on top
of the current architecture.

\paragraph{No explicit out-of-distribution detection (OOD)}

Epistemic uncertainty generally grows in regions of the data space with sparse
samples (this is an observation, not a guarantee), which means the model could
still be confidently wrong on an out-of-distribution sample. Therefore, \sptgnn{}
would need to be paired with an out-of-distribution (OOD) detection mechanism.

\paragraph{Frozen TerraMind backbone}

We keep the ViT backbone frozen during training, as propagating through a
foundation model backbone is out of our GPU budget, which caps the model's
ability to correct any modality that the fine-tune might have missed.

\paragraph{Transductive evaluation protocol}

As detailed in Section~\ref{ssec:data-splitting-strategy}, the graph topology is
built once over the full sample pool, so validation and test node features
participate in the message-passing neighbourhoods of training nodes even though
their labels are masked. The reported held-out metrics therefore reflect
\textit{transductive} performance and are best read as an upper bound on the
inductive accuracy attainable on samples from previously unseen areas.

\paragraph{Ablation coverage}

The ablations in Section~\ref{ssec:ablations} isolate the heterogeneous graph,
the MoE fusion and the ViT stream, but do not separately disable the temporal
encoder or the Moran's-$I$ auxiliary loss. Their individual contributions are
left for future work.

\paragraph{Comparison to prior geospatial GNNs}

Our headline comparison uses a tuned tabular gradient-boosting baseline. We do
not include a direct empirical comparison against the \textbf{PE-GNN} family
\citep{klemmer2023pegnn,zhao2023soc} discussed in
Section~\ref{sec:related-work}, since the published setups differ in covariates
and sampling. Benchmarking these methods on our corpus is a priority for future
work.

\section{Conclusion}\label{sec:conclusion}
\par In this study we introduced \sptgnn{}, a multi-modal, spatio-temporal
heterogeneous graph neural network for soil organic carbon (\soc{}) prediction.
Our proposed architecture which incorporates multiple types of relations between
soil samples such as position, vegetation and elevation, as well as a per-region
fine-tuned foundation model TerraMind and fusing the multiple streams of input
data (tabular, images, encoded space and time) through a cross-gated
Mixture-of-Experts (MoE). We assessed uncertainty using a $5$-member deep
ensemble coupled with post-hoc temperature scaling.

\par On our Africa sub-dataset, the deep-ensemble \sptgnn{} reaches
$R^{2}=0.762$, an RMSE of $3.51 \pm 0.48\,\mathrm{g\,kg^{-1}}$ and a MAPE of
$22.9\,\%$ on the test split, while the best single-model checkpoint reaches a
validation $R^{2}=0.864$ and MAPE of $13.7\,\%$. Ablations confirm that the
heterogeneous graph, the MoE fusion layer and the foundation-model backbone
contribute meaningfully to the model's output. At the same time, the evaluations
performed on the Europe sub-dataset, as well as Global unified dataset confirm
that the architecture transfers, though performance tends to vary with the
distribution of the training set, suggesting that source-specific variants are a
promising avenue for future work.

To our knowledge, \sptgnn{} is the first framework to combine foundation-model
feature extraction, heterogeneous relational graph attention and decomposed
uncertainty quantification for soil properties estimation. Trained models and
per-region configurations will be released through public model cards to
facilitate reproducibility and future research.

\section*{Declaration of generative AI and AI-assisted technologies in the
writing process}
During the preparation of this work the authors used Claude (Opus 4.7,
Anthropic) in order to improve the readability and language of certain sections
of the manuscript. After using this tool, the authors reviewed and edited the
content as needed and take full responsibility for the content of the
publication.

\bibliographystyle{plainnat}
\bibliography{references}

\end{document}